\documentclass{article}
\usepackage{standalone}
\input{abbrv}
\usepackage[utf8]{inputenc} % allow utf-8 input
\usepackage[T1]{fontenc}    % use 8-bit T1 fonts
\usepackage{url}            % simple URL typesetting
\usepackage{booktabs}       % professional-quality tables
\usepackage{amsfonts}       % blackboard math symbols
\usepackage{microtype}      % microtypography

\usepackage{nicefrac} 

\usepackage{amsmath}
\usepackage{amsthm}
\usepackage{amssymb}
\usepackage{thmtools,thm-restate}
\usepackage{thm-restate}
\usepackage{paralist}
\usepackage{comment}
\usepackage{dsfont}

\usepackage{bbm}
\usepackage{float}
\usepackage{balance}
\usepackage{stmaryrd}
\usepackage{multicol}

\usepackage{enumitem}
\setitemize{noitemsep,topsep=0pt,parsep=0pt,partopsep=0pt}

\usepackage{chemmacros}

\usepackage[ruled, linesnumbered]{algorithm2e} %
\usepackage{xcolor} % 
\usepackage{tikz}
\usetikzlibrary{fit,calc}
\colorlet{pink}{red!40}
\colorlet{lightblue}{blue!30}
\colorlet{lightgreen}{green!30}

\DontPrintSemicolon

\usepackage{jmlr2e}
\usepackage[margin=1in]{geometry}

\usepackage{natbib}
\setcitestyle{authoryear,round,citesep={;},aysep={,},yysep={;}}
\renewcommand{\cite}[1]{\citep{#1}}

\usepackage[vvarbb]{newtxmath}

\newcommand*{\vsepfbox}[1]{%
  \begingroup
    \sbox0{\fbox{#1}}%
    \setlength{\fboxrule}{0pt}%
    \mbox{\kern-\fboxsep\fbox{\unhbox0}\kern-\fboxsep}%
  \endgroup
}

\theoremstyle{plain} \numberwithin{equation}{section}

\numberwithin{theorem}{section}

\theoremstyle{definition}

\theoremstyle{plain}

\graphicspath{{imgs/}}
\makeatletter
\def\mathcolor#1#{\@mathcolor{#1}}
\def\@mathcolor#1#2#3{%
  \protect\leavevmode
  \begingroup
    \color#1{#2}#3%
  \endgroup
}
\makeatother
\usepackage{graphicx,wrapfig,lipsum}
\usepackage[capitalize,noabbrev]{cleveref}
\usepackage{array,multirow,booktabs}

\usepackage[
singlelinecheck=false % <-- important
]{caption}
\usepackage{makecell}

\usepackage{amsmath}
\usepackage{amssymb}
\usepackage{mathtools}
\usepackage{amsthm}
 \usepackage{url}

\begin{document}

\title{Training Characteristic Functions with Reinforcement Learning: \\ XAI-methods play Connect Four}

\author{\name Stephan Wäldchen\footnotemark[1] \email
        \href{mailto:wirth@zib.de}{waeldchen@zib.de}\\
       \addr Institute of Mathematics \& AI in Society, Science, and Technology\\
       Technische Universit\"at Berlin \& Zuse Institute Berlin\\
       Berlin, Germany\\
       \AND
       Felix Huber\footnotemark[1]
       \email \href{mailto:huber@zib.de}{huber@zib.de}\\
       \addr Institute of Mathematics \& AI in Society, Science, and Technology\\
       Technische Universit\"at Berlin \& Zuse Institute Berlin\\
       Berlin, Germany\\
       \AND
       \name Sebastian Pokutta \email \href{mailto:pokutta@zib.de}{pokutta@zib.de} \\
       \addr Institute of Mathematics \& AI in Society, Science, and Technology\\
       Technische Universit\"at Berlin \& Zuse Institute Berlin\\
       Berlin, Germany}

\maketitle

\begin{abstract}
Characteristic functions (from cooperative game theory) are able to evaluate partial inputs and form the basis for attribution methods like Shapley values. These attribution methods allow us to measure how important each input component is for the function output---one of the goals of explainable AI (XAI).
Given a standard classifier function, it is unclear how partial input should be realised.
Instead, most XAI-methods for black-box classifiers like neural networks consider counterfactual inputs that generally lie off-manifold, which makes them hard to evaluate and easy to manipulate.

We propose a setup to directly train characteristic functions in the form of neural networks to play simple two-player games. We apply this to the game of Connect Four by randomly hiding colour information from our agents during training. This has three advantages for comparing XAI-methods: It alleviates the ambiguity about how to realise partial input, makes off-manifold evaluation unnecessary and allows us to compare the methods by letting them play against each other.
\end{abstract}

\footnotetext[1]{Both authors contributed equally to this work.}

\section{Introduction}
\label{sec:introduction}

The safe deployment of AI-systems in high-stakes applications such as autonomous driving \cite{schraagen2020trusting}, medical imaging \cite{holzinger2017we} and criminal justice \cite{rudin2018optimized} requires that their decisions can be subjected to human scrutiny. The most successful models, often based on machine learning (ML) and deep neural networks (DNN), have instead grown increasingly complex and are widely regarded to operate as black-boxes.
This spawned the field of explainable AI (XAI) with the explicit aim to make ML models transparent in their reasoning. 

% European Union: \cite{goodman2017european}
\subsection{Explainable Artificial Intelligence}

%  The transparency of these systems has become a question of legality, security and ethics.
Though XAI had practical success, such as detecting biases in established data sets \cite{lapuschkin2019unmasking}, there is currently no consensus among researchers about what exactly constitutes an explainable model \cite{lipton2018mythos}.  For a good overview see \cite{adadi2018peeking}.

Models such as $k$-nearest neighbours, succinct decision trees or sparse linear models are deemed inherently interpretable \cite{arrieta2020explainable}, which makes them preferable \cite{rudin2019stop}.
However, the most impressive breakthroughs in the field of AI have only been possible with DNNs.
In this light, a second paradigm emerged: to apply these successful models and explain them post-hoc.

In this work, we focus on \emph{saliency} (or \emph{relevance}) attribution methods. Given a classifier and input, these methods rate the importance of each feature for the classifier output, often displayed visually as a heatmap, called a \emph{saliency map}. We give an overview over the proposed methods in \cref{sec:related_work}.

% on We want to focus on relevance (or salience) maps that aim to represent what part of the input the network focuses on.
% % These methods can be local or global 
% A number of these methods have sprung up, such as Zeilinger, LRP, Ribeiro (LIME) etc.

 \subsection{Characteristic Functions}\label{sec:char}
 
 Cooperative game theory considers attribution problems very similar to saliency attribution, where a common pay-off is to be fairly distributed to a number of cooperating players. In the context of ML, players correspond to features and the pay-off is the classifier score.
 Let $d\in \N$ be the number of features, and $[d] = \skl{1,\dots,d}$. One key concept is the \emph{characteristic function} $\nu\colon 2^{[d]} \rightarrow \R$ which assigns a value to every possible subset $S$ of the $d$ features, called a \emph{coalition}. We refer the reader to \cite{chalkiadakis2011computational} for a good introduction into cooperative game theory.
 
 For binary classifiers this led to the concept of \emph{prime implicant explanations} (PIE) that search for the smallest coalition $S\subset[d]$ that ensures a value of $\nu(S)=1$, see \cite{shih2018symbolic}. These explanations can be efficiently computed for certain simple classifiers like decision trees or binary decision diagrams.

 Furthermore, the \emph{Shapley values} are an established attribution method, defined as
\[
    \phi_{\nu,i} = \sum_{S \subseteq [d]\setminus\{i\}} \begin{pmatrix} d-1 \\ \bkl{S} \end{pmatrix}^{-1} \kl{ \nu(S \cup \{i\}) - \nu(S) }.
\]
Shapley values are the unique attribution methods that satisfy certain desirable fairness criteria, see \cite{shapley201617} and \cref{apx:tlight}.

Both PIE and Shapley values are thus defined over desirable properties, and the actual algorithm to compute them depends on the model. In contrast, most saliency attribution methods for neural networks are defined directly over algorithmic instructions and lack definite properties that make them useful\footnote{Except the gradient maps, though these suffer from other shortcomings as explained in \cref{apx:off_manifold}}. Even the saliency methods that are inspired by PIE and Shapley values (see \cref{sec:related_work}) do not maintain their desirable theoretical properties while at the same being computationally efficient \cite{macdonald2020explaining}. This means that the only way to judge whether these saliency methods have merit is to evaluate them in practical scenarios.

\subsection{Evaluating Saliency Methods}

 \begin{figure*}[t]
    \centering
    \includegraphics[width=0.9\textwidth]{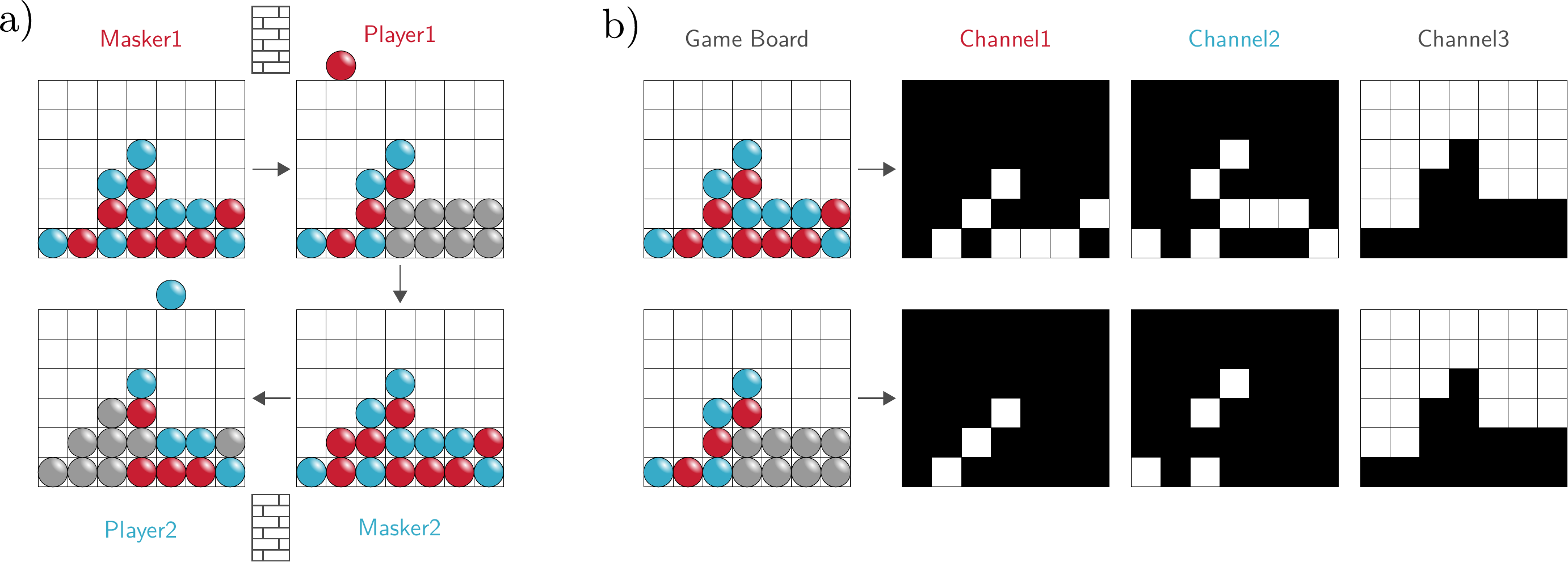}
    \caption{Connect 4 with missing colour information shown in grey. a) A game between two pairs of maskers and players. The maskers decide which colour information to pass on to the players with an upper limit of revealing half the played pieces. The player has to decide his move based on the information sent by the masker.
    Our the players are both represented by the same policy network. Which information to reveal will later be chosen by different saliency methods.
    b) The game board is encoded in three input channels: Two binary matrices indicating the pieces of each player and one indicate the open fields.
    %\footnote{We indicate open fields instead of occupied ones because it works better with the zero-padding in the convolutional layers that would otherwise suggest a larger game board.}
    With full information the encoding is redundant, with only partial information the agent is unsure about the colour of some pieces but can still make valid moves.
    }
    \label{fig:highlevel}
 \end{figure*}

In \cite{doshi2017towards} the authors differentiate between \emph{human-based} and \emph{functionally-grounded evaluation}. The former has the advantage of measuring directly what we want, namely that explanations are legible and helpful to humans. On the other hand, human-based evaluations are costly and hard to generalise from one task to another. They also cannot discern between a network with unintuitive reasoning and a saliency method that produces unintuitive results.
Functionally grounded evaluation aims to design proxy tasks whose success is correlated with the quality of the explanation. They often ask, which information is useful to the interpreted network itself, see \cite{mohseni2021multidisciplinary}, so that only the quality of the saliency method matters.
 This approach also allows for much larger scale experiments. However, not all proxy-task are necessarily linked to good explanations \cite{biessmann2021quality}.

The proxy task we consider here is: \emph{successful play of abstract games with limited information.} In our case study we investigate the game \emph{Connect Four} \cite{allis1988knowledge}.
We make use of the fact that neural networks have emerged as one of the strongest models for reinforcement learning, and e.g. constitute the first human competitive models for Go \cite{silver2017mastering} and Atari games \cite{mnih2015human}.

Our exact setup is illustrated in \cref{fig:highlevel}. 
A \emph{masker} and a \emph{player} are paired against a second team of the same form. The masker, presents a limited amount of information (colour features) to their player, who then selects the next move. The full board state is then given to the masker of the opposing team who selects the information for their player, and so on until the game is finished either by one party winning or a draw.
The players are modelled by DNNs without memory and base their decision only on the information currently provided by the masker. During training, the masker will select information randomly up to a varying maximal amount. For the comparison of the saliency methods, the masker will instead be represented by a method that explains the move that the player would have made given full information. The most salient features are then given to the player for their actual move.

\subsection{Our Contribution}

First, we give an overview over the existing saliency methods and explain how they are vulnerable to manipulation if they rely on off-manifold inputs.
We show that for image obfuscation, one of the most used evaluation metrics, the best-performing methods give artefactual explanations that rely on ``superstimuli'', a phenomenon resulting from evaluating classifiers off-manifold.

As a remedy, we directly train agents as characteristic functions for reinforcement learning, by randomly hiding colour features and show that this setup delivers results comparable to training solely on full information.
Additionally, we demonstrate a relatively monotonous relationship between information and performance in the game, which justifies the setup explained in \cref{fig:highlevel} as a sensible proxy task for XAI methods.
Since our agents can handle partial input, we can directly compute Shapley values via sampling. These are theoretically well understood and rely only on counterfactual input the agent has been trained on---which makes this sound attribution method. We demonstrate their usefulness by comparing to the ground truth available for certain board situations. Additionally, we learn that training on hidden input is not enough to ensure interpretability if the training is linked to the wrong objective.

We then compare a selection of XAI methods in a round-robin tournament in Connect Four, which has some advantages that image obfuscation comparisons generally lack:\vspace{-0.2cm}
\begin{enumerate}
\itemsep-0.1em 
    \item It is canonically clear how missing information should be modelled (since it is included in the training).
    \item There is no need to evaluate the classifier off-manifold. \item We have a concrete task (Winning the game) to compare the XAI-methods.
\end{enumerate}

\section{Related Work}\label{sec:related_work}

A lot of work has been done to design methods that produce saliency maps for neural networks. So far, these methods are not equipped with theoretical guarantees that they fulfil a certain quality property, as compared to Shapley values and PIE, see  \cref{sec:char}. They are instead motivated by heuristic arguments and then numerically evaluated. We give a short overview over the existing heuristics and then explain that first, methods relying on off-manifold counterfactuals are manipulable and secondly, evaluations relying on off-manifold inputs are manipulable as well. 

\subsection{Saliency Methods}

We restrict our analysis to local, post-hoc saliency methods for neural network classifiers, both model specific and model-agnostic. We differentiate the following three categories.

\paragraph{Local Linearisation} Linear methods are considered interpretable, so it is a natural approach for nonlinear models to instead interpret a local linearisation. In this category we find gradient maps \cite{simonyan2013deep}, SmoothGrad \cite{smilkov2017smoothgrad}, which samples the gradient around the input, and LIME \cite{ribeiro2016should} which applies the classifier to samples around the input then fits a new linear classifier to the resulting input-label-pairs.

\paragraph{Heuristic Backpropagation}

These methods replace the chain-rule of gradient backpropagation with different heuristically motivated rules and propagate relevance scores back to the input. One of the earliest methods methods for neural networks was deconvolution \cite{zeiler2014visualizing}, newer methods include GuidedBackpropagation (GB) \cite{springenberg2014guidedbackprop}, DeepLift \cite{shrikumar2017learning}, DeepShap \cite{NIPS2017_7062shap} and LRP \cite{bach-plos15}. These methods have the advantage of being computationally very fast and applicable in real-time.

\paragraph{Partial Input}

These methods rely on turning a classifier function $f$ and an input $\bfx$ into a characteristic function $\nu_{f, \bfx}$. The standard way to define $\nu_{f, \bfx}(S)$ for a feature set $S$, put forth by \cite{lundberg2017unified}, is to regard the missing features $\bfx_{S^c}$ as random variables and take an expectation value conditioned on the given parameters $\bfx_{S}$, i.e. 
\begin{equation}\label{eq:class_to_char}
    \nu_{f,\bfx}(S) = \E_{\bfy}\ekl{f(\bfy)\,|\, \bfy_S = \bfx_S } = \int f(\bfx) p(\bfx_{S^c} \,|\, \bfx_S) \,\text{d}\bfx_{S^c}.
\end{equation}
 Being able to evaluate partial input, these methods either optimise an objective similar to prime implicant explanations, for example \emph{Rate-Distortion Explanations} (RDE) \cite{macdonald2020explaining} and \emph{Anchors} \cite{ribeiro2018anchors}, or approximate Shapley values \cite{sundararajan2020many}.
 This however is computationally infeasible to do exactly, so these methods instead rely on heuristic strategies that do not carry over the quality properties of PIE and Shapley values.
 Additionally, the soundness of these heuristics depends strongly on how correctly the conditional distribution $p(\bfx_{S^c} \,|\, \bfx_S)$ is modelled, as we will discuss next.

\subsection{Off-Manifold Input}

\begin{figure}[t]
    \centering
    
  \centering\scriptsize
  \begin{tabular}{@{}c@{\;}c@{\;}c@{\;}c@{\;}c@{\;}c@{}}
  
    Image & FW & AFW & LCG & LAFW & Sensitivity\\
    \includegraphics[height=2.2cm]{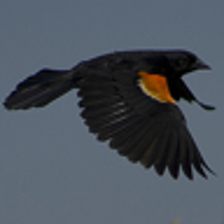} &
    \includegraphics[height=2.2cm]{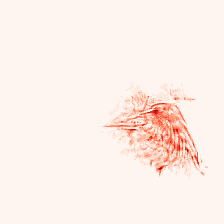} &
    \includegraphics[height=2.2cm]{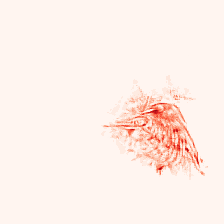} &
    \includegraphics[height=2.2cm]{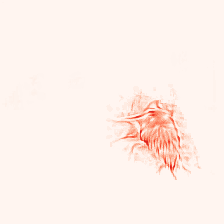} &
    \includegraphics[height=2.2cm]{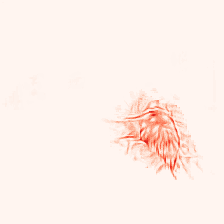} &
    \includegraphics[height=2.2cm]{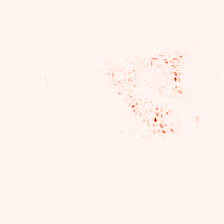}
  \end{tabular}

    \caption{\small RDE-Explanations of a bird image taken from \cite{macdonald2021interpretable} with permission of the authors. The proposed optimisation methods (FW, AFW, LCG, LAFW) search for the smallest set of pixels that still maintain the classification of ``bird'', if the other pixels are randomised. All produce a mask that creates a new bird head as a ``superstimulus'' mask. The sensitivity map does not show this behaviour.
    The distribution of the randomised pixels is independent from the selected set, which means the contours of the set will be visible and detectable by the pattern matching network. Modelling the noise after true distribution $p(\bfx_{S^c} \,|\, \bfx_S)$ (as in \cref{eq:class_to_char}) 
    would prevent this effect, since a monochrome selection of black pixels would likely be inpainted with black as well. Then no bird's head would appear.}
    \label{fig:stimulus}
\end{figure}

These post-hoc relevance methods have in common that they consider counterfactual information: \emph{``What if I would change this part of the input?''}. We explain how this applies to each method in \cref{apx:off_manifold}.

For this reason, the saliency methods can all be manipulated by principally the same idea: replace an existing model by a another one that agrees on the data manifold but not off-manifold, where the behaviour can be chosen arbitrarily. This allows to hide biases in classifiers for \emph{on-manifold} inputs almost at will, as demonstrated for gradient maps and integrated gradients \cite{anders2020fairwashing, dimanov2020you}, LRP \cite{anders2020fairwashing, dombrowski2019explanations}, LIME \cite{slack2020fooling, dimanov2020you}, DeepShap \cite{slack2020fooling, dimanov2020you}, Grad-Cam \cite{heo2019fooling}, Shapley-based\cite{frye2020shapley} and general counterfactual explanations \cite{slack2021counterfactual}.

\paragraph{Image Obfuscation} In the absence of human annotations (such as bounding boxes or pixel-wise annotations), some functionally grounded evaluation for image data obfuscate part of the image and measure how much the classifier output changes \cite{mohseni2021multidisciplinary}. The idea is this: keeping the relevant features intact should leave the classification stable, obfuscating them should rapidly decay the classifier score.
This method was introduced as \emph{pixel-flipping} \cite{WojBGM2017pixelflip} and used to evaluate XAI-methods for image recognition \cite{fong2017interpretable, macdonald2019rate} and Atari games \cite{huber2021benchmarking}.

In \cite{macdonald2019rate} the authors directly optimise for a mask that selects sparse features which maintain the classifier decision, severely outperforming competitors. In \cite{macdonald2021interpretable} this method gets improved further with methods from convex optimisation (Frank-Wolfe optimiser). However, visually inspecting the produced saliency maps reveals they create new features that were not present in the original image. This is possible over a mechanism similar to adversarial examples, see \cref{fig:stimulus}, which means: \emph{the optimal mask creates its own features!}
Since the distribution of the obfuscated part of the image is independent from the rest the created images lie off-manifold which makes this effect possible. To keep the image on-manifold, the obfuscated part would have to be sampled from the true conditional data distribution $p(\bfx_{S^c} \,|\, \bfx_S)$ which is not known.
This shows that proxy tasks have to be designed with care if they are meant to be useful for comparing saliency methods.
Not only are the methods vulnerable to manipulation by going off-manifold, the evaluation tasks themselves can be exploited by making use of off-manifold inputs.

\section{Setup}\label{sec:setup}

Instead of trying to turn a classifier function into a characteristic function, we propose to directly train on hidden features.
 Policy and value functions (see \cite{li2017deep}) for agents that play simple two-player, turn-based games (such as Go, Checkers, Hex) are particularly well suited for this task.
 
 The logic of these games is complex enough to make the use of black-box functions such as neural network sensible, and in the case of Go the unbeaten standard.
 At the same time, the input is low-dimensional and discrete.
 Additionally, the input components have weaker correlations between neighbours, i.e. a random configuration of Connect Four pieces can still be a valid game\footnote{if the number of pieces for each player is balanced}, whereas a random assortment of pixels will almost surely not be a valid image. These factors facilitate sampling partial inputs during training. \footnote{For image data by comparison, simply hiding random pixels would not do, since the image can likely still be inferred by inpainting.}

  \begin{figure}[t]
     \centering
     \includegraphics[width=0.7\textwidth]{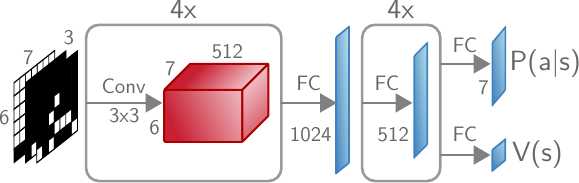}
     \caption{Network architecture for the PPO-agent. The convolutional layers (Conv) have $3\times 3$ filters, 512 channels, $1\times 1$ stride and use zero-padding. The fully connected (FC) layers for the policy and value head use softmax and tanh-activation respectively. All other layers use ReLU-activations.}
     \label{fig:architecture}
 \end{figure}

 \subsection{Hiding the Player Colour}
 
 The most straight-forward way of hiding information from an agent would be to complete hide a game field. However, we want to preserve the ability to select legal moves, and let the sensibility of the move be the only concern of the agent (instead of legality).
 For a lot of the games (e.g. Connect Four, Go, Hex) knowing which fields are occupied allows to make valid moves. For others (Chess, Checkers) valid moves depend on the colour and type of the pieces, so we will concentrate on the former.
 
 To hide the colour information, we represent a game position as three binary matrices indicating which fields are occupied by the first player (red), the second player (blue) and which remain free. The colour information can be hidden by setting the entries in the respective matrix to zero. We illustrate this concept in \cref{fig:highlevel} a) for the game of Connect Four.

\subsection{Reinforcement Learning for Connect Four}
 
 The game of Connect Four was chosen because of its simplicity and low input dimension.
 Deep-RL has been applied to Connect Four in the form of both Deep-Q-Learning \cite{dabas2022solving}, Policy Gradient (PG) \cite{crespo2020master} and AlphaZero-like approaches \cite{wang2021adaptive, clausen2021improvements}.
 The AlphaZero-like approach combines a neural network with policy and value head with an MCTS\footnote{Monte-Carlo Tree Search, see \cite{silver2017mastering}} lookahead to make its decision. Even though it has emerged as the most powerful method, we want to explain purely the network decision without the MCTS involved. We thus follow the PPO approach of \citeauthor{schulman2017proximal}, since it yields a strongly performing pure network-agent for Connect Four even without the help of the MCTS \cite{crespo2020master}.
 
 \paragraph{Our PPO-agent} We represent the input by a state $\bfs \in \skl{0,1}^{3\times6\times7}$ with three channels as explained in \cref{fig:highlevel} (b). The value function that predicts the expected reward $V(\bfs)$ and the policy function that determines the probability $P(a\,|\,\bfs)$ of taking action $a\in[7]$ are both modelled by the same network with with a policy and a value head, see \cref{fig:architecture}.  
 Our architecture extends the model of \citeauthor{crespo2020master} by two fully connected layers, which empirically yields better performance. Agents can play \emph{competitively}, always choosing the most likely action $a^* = \argmax P(\cdot\,|\,\bfs)$ or \emph{non-competitively}, sampling from $P(\cdot\,|\,\bfs)$. We give a full overview over architecture and training parameters in \cref{apx:training}.

  \paragraph{Hiding features during training} During self-play, we randomly hide the colour feature of a certain percentage of fields by setting the respective entry in the the first and second input channel to zero. The information that the field is occupied remains in the third channel. Every turn $p_h$ is drawn uniformly from $[0,p_h^{\text{max}}]$. Let $t\in [42]$ be the turn number, then we hide the colour information of $\floor{p_h t}$ pieces selected uniformly at random.
  We explored different values for $p_h^{\text{max}}$ and trained the following agents:
  \begin{itemize}
     \item FI: PPO-Agent trained with full information,
     \item PI-50: Partial information with $p_h \sim \CU([0,0.5])$,
     \item PI-100: Partial information with $p_h \sim \CU([0,1])$.
 \end{itemize}
  
    \begin{table}[t]
\centering
a) Win Rate against MCTS \\[0.2cm]
 {\small
 \begin{tabular}{l|c|c|c|c}
            & Orig. & FI & PI-50 & PI-100 \\\hline 
      MCTS 500 & 0.92 & \textbf{0.972} & 0.793 & 0.684\\
      MCTS 1000 & 0.896 & \textbf{0.936} & 0.66 & 0.469\\
      MCTS 2000 & 0.825 & \textbf{0.91} & 0.497 & 0.328\\
 \end{tabular}
 }\\[0.3cm]
b) Number of Optimal Moves \\[0.2cm]
 {\small
 \begin{tabular}{c|c|c|c|c}
      Agent & Orig. & FI & PI-50 & PI-100 \\ \hline
      Correct Moves & 38 & \textbf{39} & \textbf{39} & 29
 \end{tabular}
 }
 \caption{Comparison of our agent with the original proposal by \cite{crespo2020master} in competitive mode. The numbers show the winrate against the MCTS with different simulation limits over 1000 games. b) For a game between two Connect Four solvers, we tracked how many of the optimal moves were correctly predicted by the different agents, i.e. were given the highest probability in the policy output. The optimal game always takes 41 moves in total.}
 \label{tab:comparison}
 \end{table}

  \subsection{Benchmarking} \label{sec:benchmarking}
 
 To demonstrate that this setup still trains capable agents we compare them to the benchmark results from the original setup presented in \cite{crespo2020master}. We let our agents play in competitive mode against an MCTS-agent taken from \cite{cpp_mcts}. For three different difficulties, the MCTS is allowed to simulate 500, 1000 or 2000 games. Additionally, we used a game played by two perfect solvers\footnote{Taken from \cite{perfect_agent}} and measured how many of the 41 moves were predicted correctly by our agents.  
 For the results see \cref{tab:comparison}. 
 Our FI-agents performs best, both against the MCTS and predicting the optimal moves. Incorporating partial information into the training leads to a worse performance by PI-50 and PI-100. Nevertheless, at least the PI-50 is a capable agent that could not be beaten by the authors. We thus opt to use the PI-50 agent to compare the saliency methods in \cref{sec:tournament}.

\begin{figure}[t]
    \centering
    
  \centering
  \begin{tabular}{@{}c@{\;}c@{\;}c@{}}
    \hspace{0.31cm}\includegraphics[height=3.5cm]{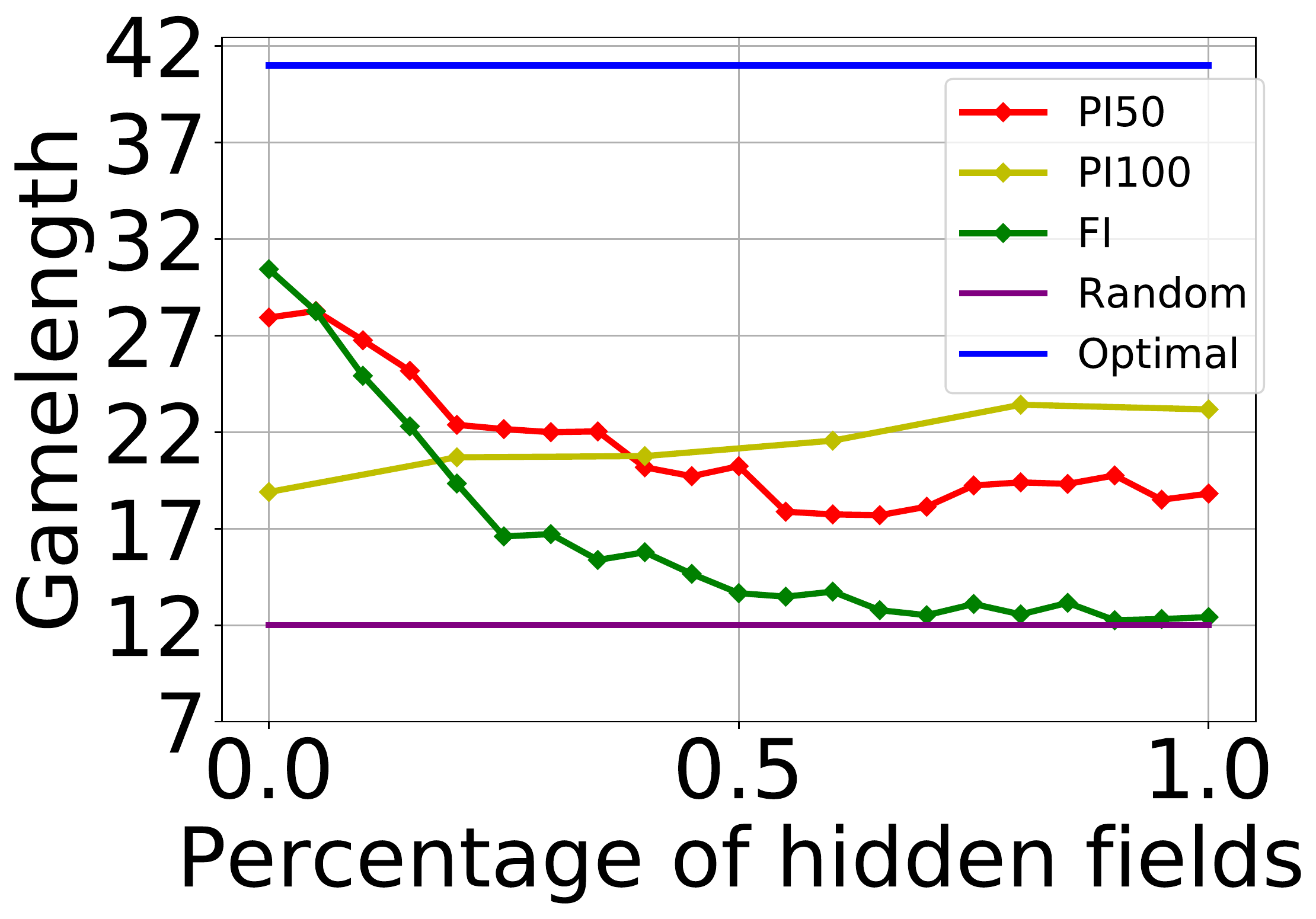}
    & 
    \includegraphics[height=3.5cm]{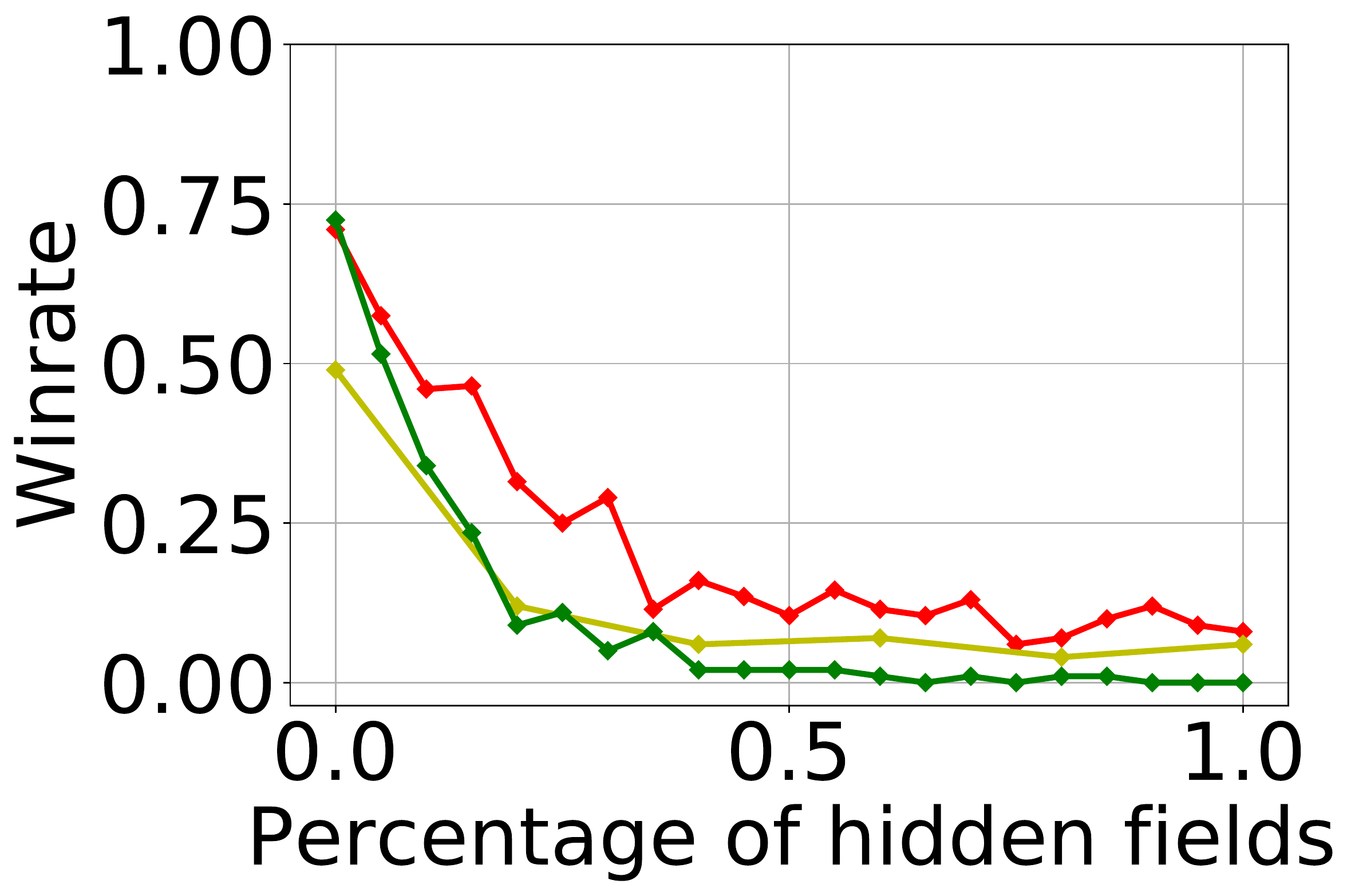}
    &
    \includegraphics[height=3.5cm]{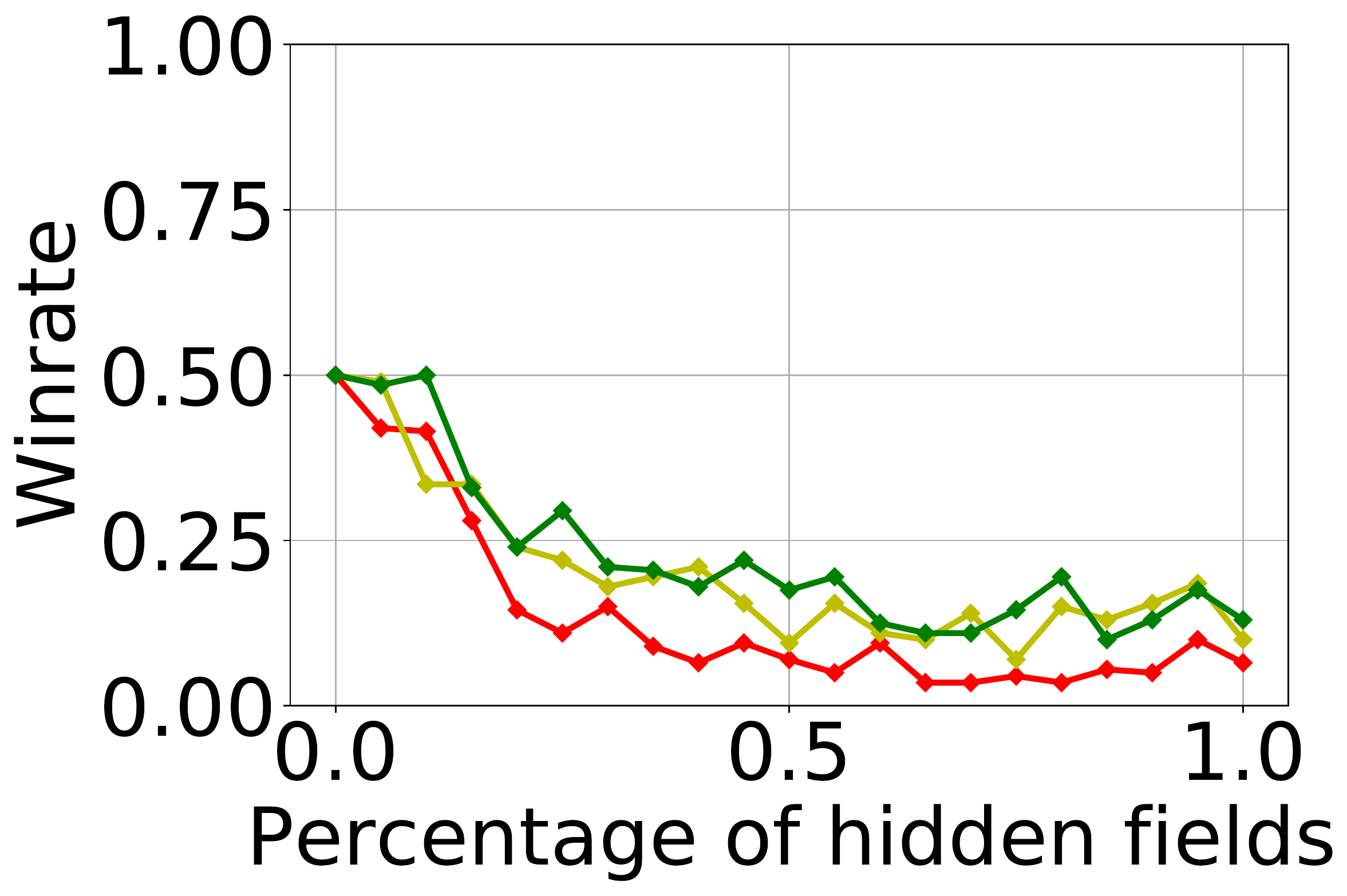} \\
    a) & b) & c)
  \end{tabular}

    \caption{Relationship between the percentage of hidden fields and game performance, measured in game length against a perfect solver (a), winrate against an MCTS1000 (b) and against the agent itself with full information (c) for the FI, PI-50 and PI-100 agents. In (a) we give as reference a random agent who plays an average of 12 turns before losing and another optimal solver who always plays 41 turns. Our agents never win against the solver. The FI-agent start out strongest, but was never trained on hidden features, so drops towards the random agent. The PI-50 starts with slightly weaker game length and but keeps an advantage over random even for no colour information at all. The PI-100 agents shows a weakest performance given full information but even rises slight with less information. Every game takes at least 7 turns and at most 42. In (b) and (c) all agents decrease in performance with fewer features revealed.}
    \label{fig:info_performance}
    \vspace{-0.4cm}
\end{figure}

 To show that for each agent more information is indeed useful\footnote{This is not obvious: during training the agent could get used to an average amount of partial information and play worse if given more colour features.}, we tracked their performance playing non-competitively for different amounts of randomly hidden colour features, see \cref{fig:info_performance}. For the FI-agent and the PI-50 agent we see near-monotonous decay in game performance against the optimal agent, an MCTS-1000 and themselves with full information. This justifies our idea of a proxy task to compare saliency methods: select the most useful 50\% of features that allow the PI-50 agent to win the game!

\section{Explanations with Partial Information}\label{sec:tlight}

Since our agents have been trained with missing information, their policy and value functions can be seen as a characteristic function with respect to the colour features. For $t\in [42]$ let $\bfs_t \in [0,1]^{3\times 6 \times 7}$ be a board state after turn $t$ and $S \subseteq [t]$ be a set of colour features (of a subset of the $t$ played pieces). Then we define $\bfs^{(S)}$ as a partial board state including only the colour features in $S$, as in \cref{fig:highlevel} (b), second row.

Let $a^* = \argmax P(\,\cdot\,; \bfs)$, then we can define $\nu^{\text{pol}}: 2^{[t]} \rightarrow [0,1]$ and $\nu^{\text{val}}: 2^{[t]} \rightarrow [-1,1]$ as
\[
 \nu^{\text{pol}}(S) = P(a^*; \bfs^{(S)}) \quad\text{and}\quad
 \nu^{\text{val}}(S) = V(\bfs^{(S)})
\]
and interpret them as the characteristic functions from the policy and value output for a given board state. From now on, we use the characteristic function associated with the \emph{policy} network, since this is the part that actually plays the games, whereas the value network is only involved in training.
This allows us to directly compute explanations for them in the form of Shapley values or prime implicant explanations. We now explain how we efficiently approximate both.

\subsection{Sampling Shapley Values}

Computing Shapley values is $\textsf{\#P}$-complete \cite{deng1994complexity} , but they can be efficiently approximated by sampling. The simplest approach is to utilise the fact that the Shapley values for $t$ features can be rewritten as
\[
 \phi_i = \frac{1}{t!} \sum_{\pi \in \Pi([t])} \kl{\nu(P^{\pi}_i \cup  \skl{i}) - \nu(P^{\pi}_i)},
\]
where $\Pi([t])$ is the set of all permutations of $[t]$ and $P_i^{\pi}$ the set of all features that precede $i$ in the order $\pi$. To approximate the whole sum we can sample uniformly from $\Pi([t])$. To stick true to our philosophy of evaluating $\nu$ only on-manifold, we can only use PI-100 to calculate the Shapley-values who has been trained on all levels of hidden features. To use the PI-50 agent, we define \emph{partial} Shapley-values for a hidden percentage $p_h$ by only sampling from 
\[
\Pi_i^{p_h} = \skl{\pi \in \Pi([t]) \;\text{s.t.}\; \bkl{P_i^{\pi}} \geq p_ht},
\]
which are all permutations that have input $i$ in the last $p_ht$ position. Sampling from $\Pi_i^{0.5}$ makes sure that at least 50\% of colour information is disclosed. We show in \cref{apx:partial} that we keep the symmetry, linearity and null player criterion, but loose efficiency.

To get an $(\epsilon, \delta)$-approximation $\bar{\phi}_i$ of the true Shapley values $\phi_i$ in the sense that
$
  \P\ekl{\bkl{\phi_i - \bar{\phi}_i} \leq \epsilon} \geq 1 - \delta
$, we need
$
 N_{\epsilon,\delta} = \frac{1}{2} \epsilon^{-2} \log(2 \delta^{-1})
$
many samples, according to the worst-case bound given by the Hoeffdings-inequality \cite{hoeffding1994probability}. For our comparison we choose a $(0.01,0.01)$-approximation which amounts to $\approx26500$ samples.

More efficient methods to sample Shapley-values have been developed utilising group testing \cite{jia2019towards} or kernel-herding \cite{mitchell2021sampling} both providing a quadratic improvement of the accuracy in terms of number of function evaluations although with some computational overhead.
We stick, however, to this simple approach since our input dimension is small and there is significant overhead in both the kernel-herding and group testing.

\subsection{Prime Implicant Explanations with Frank-Wolfe}

Prime implicant explanations can be efficiently calculated for simple classifiers like decision trees or ordered binary decision diagrams \cite{shih2018symbolic}. In \cite{macdonald2020explaining} the authors extended the definition of prime implicants to a continuous probabilistic setting to explain neural networks, although the authors also showed that is $\SNP$-hard to find them even approximately for general networks. To solve the problem heuristically, they optimise for implicants of size $k\in [t]$ via convex relaxation, which forms the basis for RDE. Adapted to our scenario the corresponding objective becomes 
\[
 S^* = \argmin_{\bkl{S} \leq k } (\nu([t]) - \nu(S))^{2}.
\]
Whereas \citeauthor{macdonald2020explaining} rely on an approximation to \cref{eq:class_to_char}, our formulation has no probabilistic aspect, since we directly access a characteristic function. In \cite{macdonald2021interpretable} the authors show how to minimise this functional efficiently with Frank-Wolfe solvers, a projection-free method for optimisation on convex domains \cite{pokutta2020deep}. We copy their approach and apply it to find small prime implicant explanations for varying $k$, a saliency attribution which we call the \emph{FW-method}. A detailed description of the FW-method is included in \cref{apx:FrankWolfe}.
The Frank-Wolfe method might attribute the maximum weight of 1 to multiple features, so we break ties randomly when selecting the most salient ones.

 \begin{figure}[t]
    \centering
    
  \centering
  \begin{tabular}{@{}c@{\;}c@{\;}c@{}}
  
     & Shapley Sampling & FW \\
    \includegraphics[height=3.8cm]{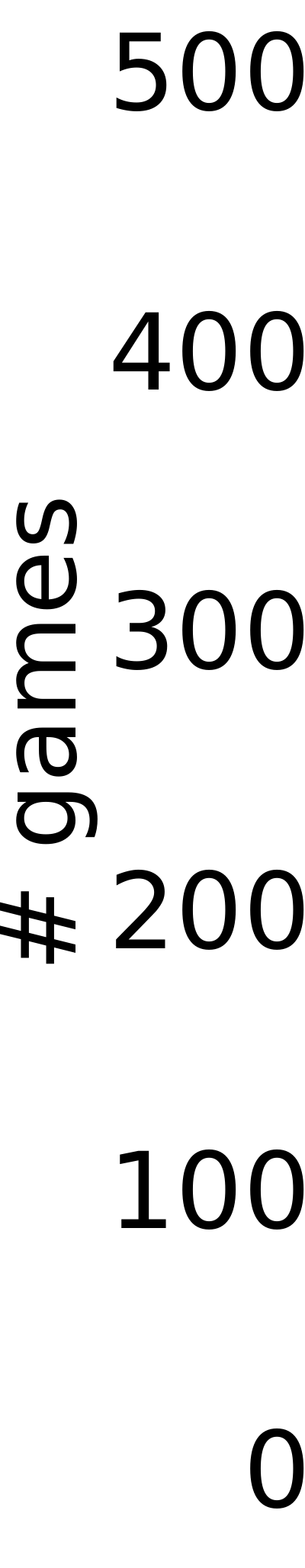} &
    \includegraphics[height=3.8cm]{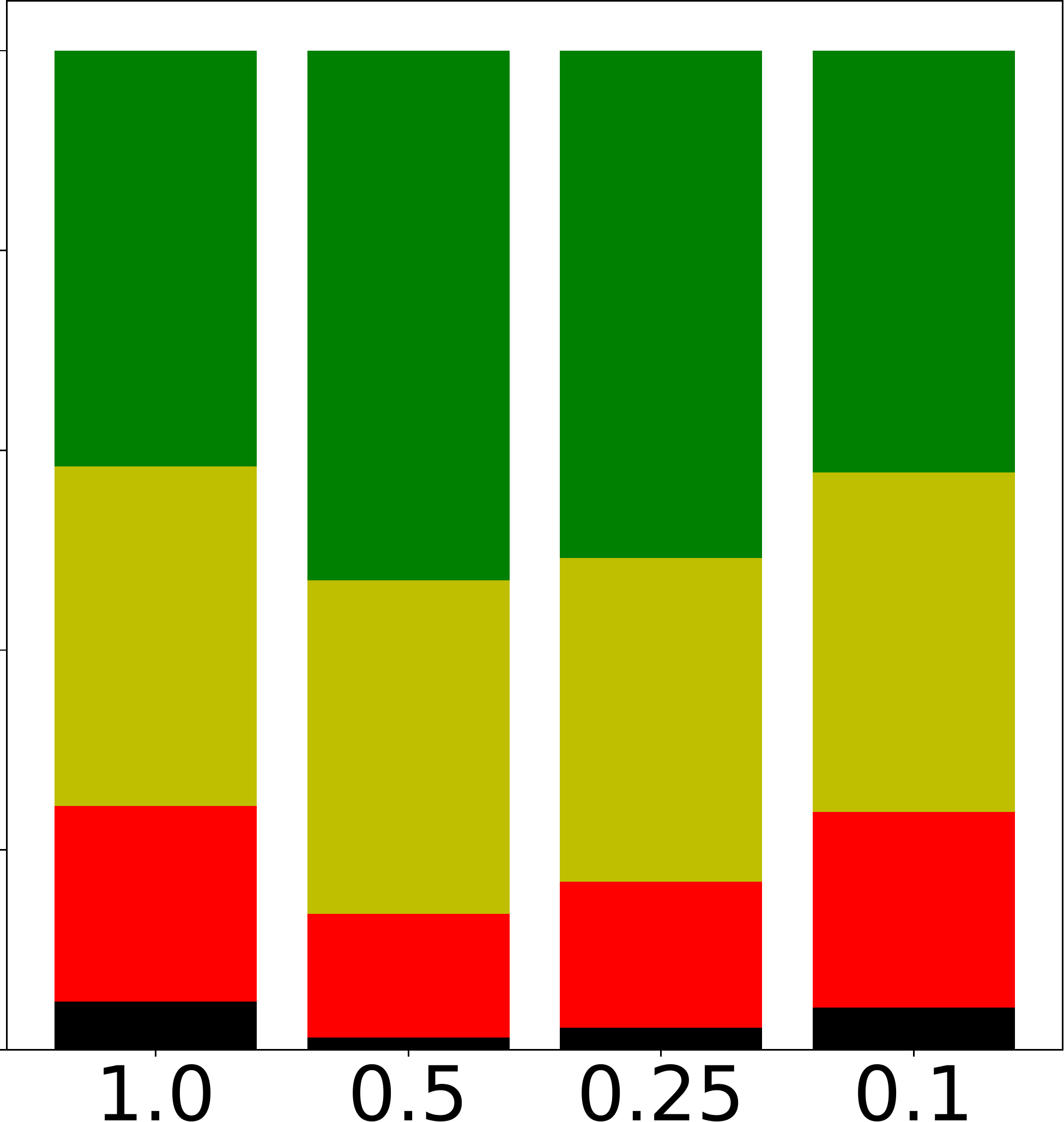} &
    \includegraphics[height=3.8cm]{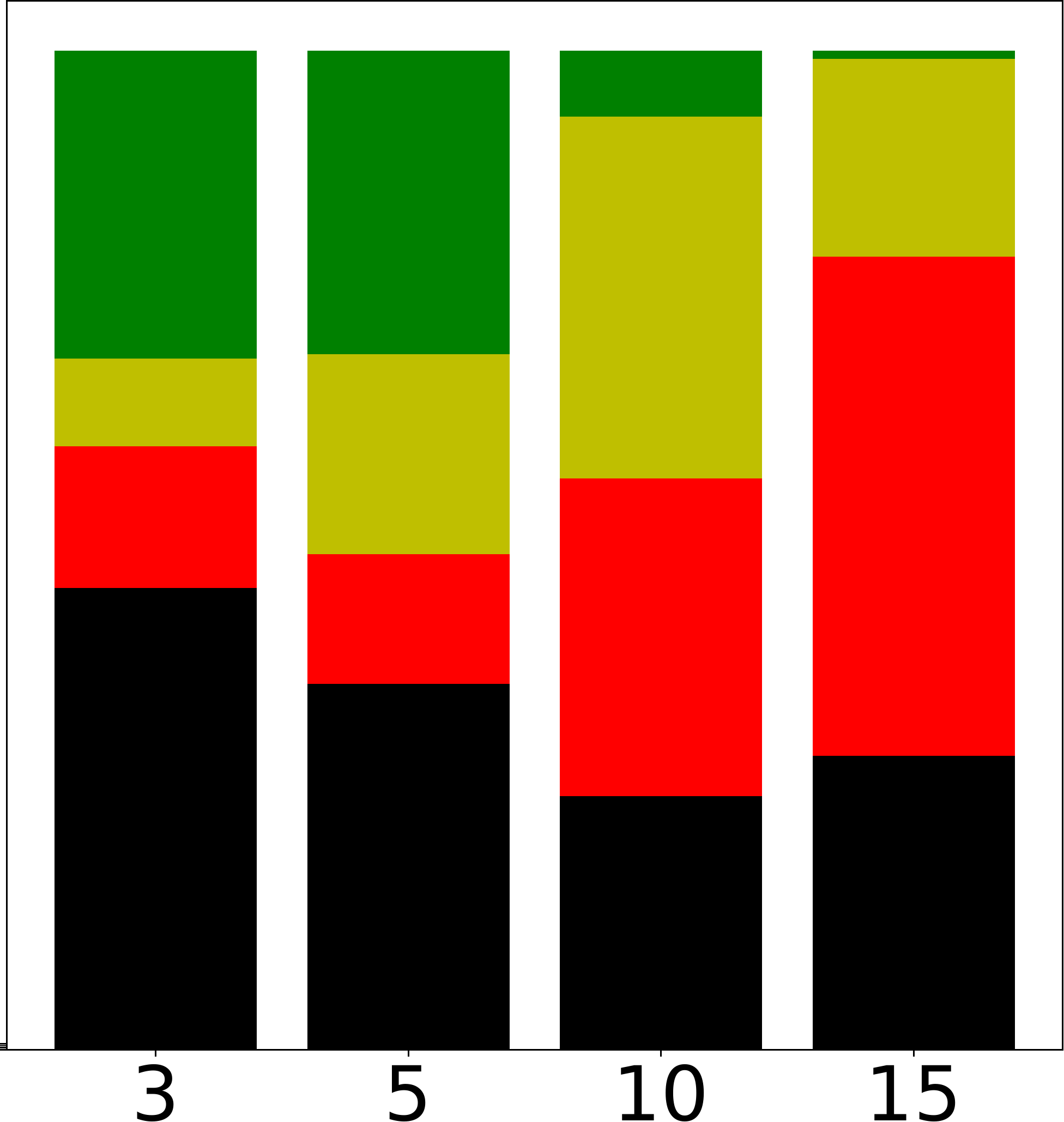}  \\[-2pt]
    & $p_h$ & $k$
  \end{tabular}

    \caption{Comparison of the Frank-Wolfe-based and Shapley Sampling-based attribution methods with the ground truth. The bars show how often the method identified the 3 (green), 2 (yellow), 1 (red) or 0 (black) of the three most important game pieces. We observe that Shapley sampling is able to identify the most pieces correctly for a hidden percentage of $p_h=0.5$. The Frank-Wolfe method performs generally worse than Shapley Sampling. Smaller $k$ tend to polarise the results with more boards where either all three or zero of the correct pieces have been found. }
    \label{fig:tlight}
\end{figure}

\subsection{Finding Ground Truth Pieces}\label{sec:ground_truth}

The policy network of the PI-agent is able to find a winning move in 99\% of cases. It stands to reason that the three pieces that are completed by the move can be seen as a \emph{ground truth} features for the decision.

We use this compare partial Shapley Values with the FW-method. We let the agent play against itself and registered 500 final board states for which the agent was at least 90\% sure of the winning move. All game pieces that form a line of four with the winning move\footnote{It can happen that one move completes multiple lines.} are considered as ground truth for the focus of the policy network.
We track how often the three most salient pieces according to the attribution methods are among the ground truth pieces. For 500 games we note whether three, two, one or zero pieces are correct, see \cref{fig:tlight}, for partial Shapley values with varying $p_h$ and the FW-method for varying $k$.

\paragraph{Shapley Sampling} Partial Shapley values generally yield good results, finding at least two correct pieces in at least 75\% of all boards. However, calculating the Shapley-values all the way up to $p_h=1$ gives worse results than stopping at 0.5, both for PI-50 and PI-100. The plots for PI-100 can be found in \cref{apx:tlight}. Further investigating revealed that the output of the PI-100 policy function for $p_h>0.5$ becomes essentially random for many board situations. A possible reason is that for low information the optimal move becomes ill-defined, and the entropy loss is to be too small to force the policy to be approximately uniform. We will come back to this point in \cref{sec:conclusion}. 

\paragraph{Frank Wolfe Solver} The FW method is faster than Shapley sampling, but exhibits worse performance. Interestingly, for small $k$ it polarises, having a higher percentage of boards where it finds either all or non of the ground truth pieces. For large $k$ it usually converges to an attribution that selects many pieces with maximum value of 1. Breaking ties randomly then leads to average results. For small $k$ it oftentimes selects pieces that suggest the right move but for a different reason than the ground truth pieces.
Additionally, the FW performance suffers from relying on convex relaxation of set membership, which means optimising over continuous colour features, thus going off-manifold. This can be seen by the fact that the continuous input values almost always lead to the right policy, but selecting the $k$ most salient features, thus thresholding the input back to binary values, sometimes leads to a different policy.

\section{Comparing XAI-methods}\label{sec:tournament}

 \begin{figure*}[t]

  \centering\scriptsize
  \begin{tabular}{@{}c@{\;}c@{\;}c@{\;}c@{\;}c@{\;}c@{\;}c@{\;}c@{\;}m{0.6cm}}
  \includegraphics[width=0.112\textwidth]{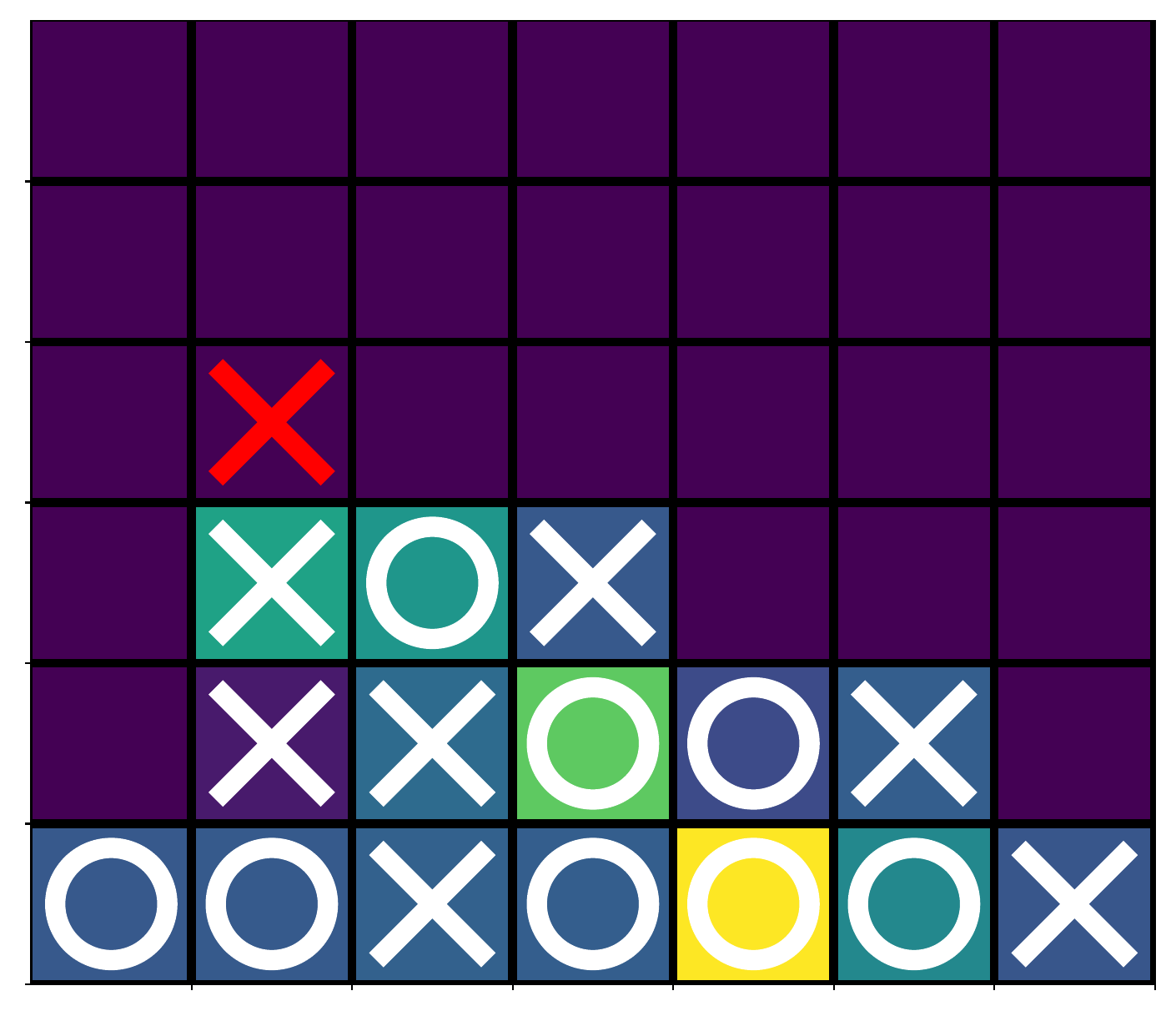}
        &
        \includegraphics[width=0.112\textwidth]{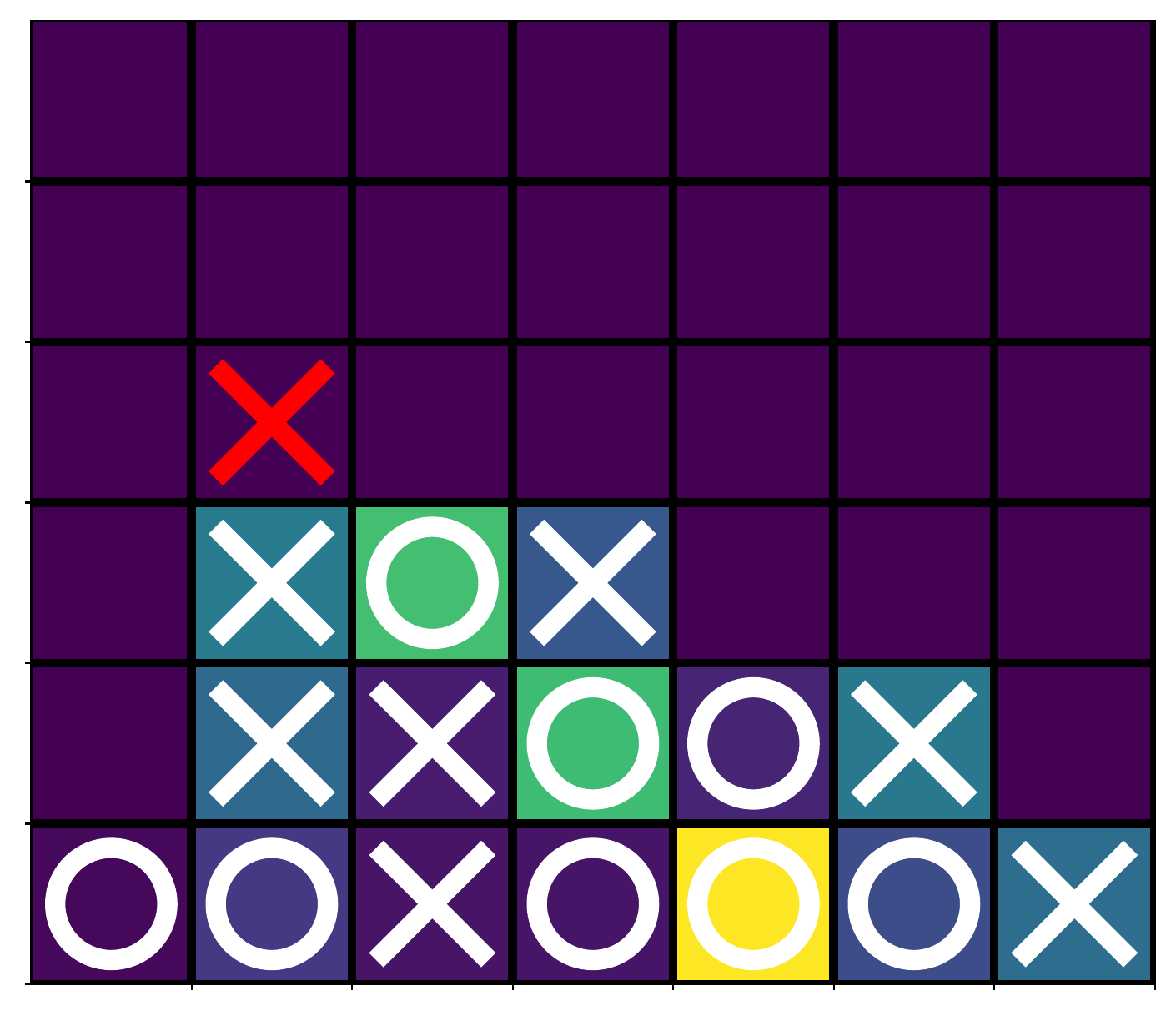}
        &
        \includegraphics[width=0.112\textwidth]{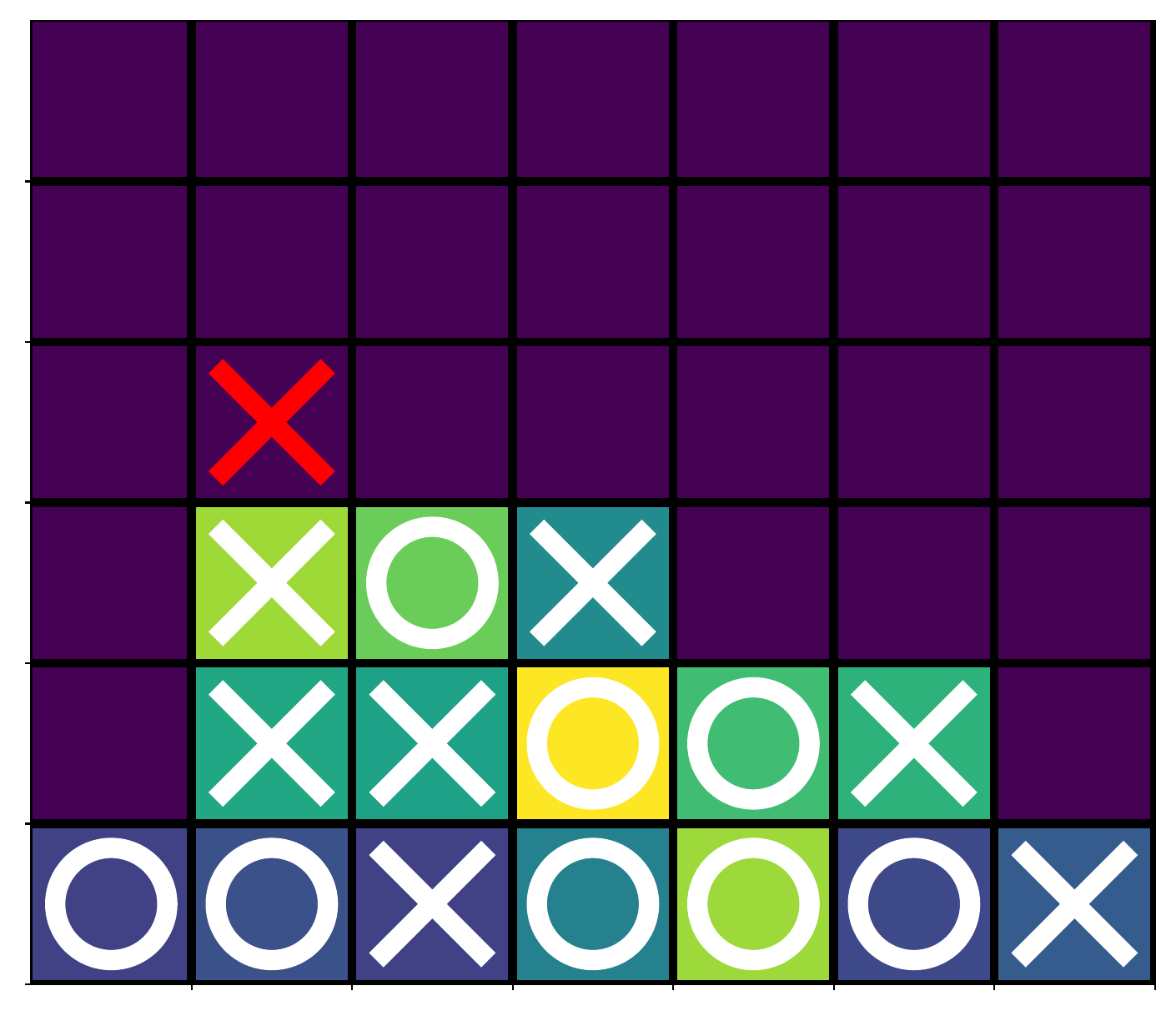}
        &
        \includegraphics[width=0.112\textwidth]{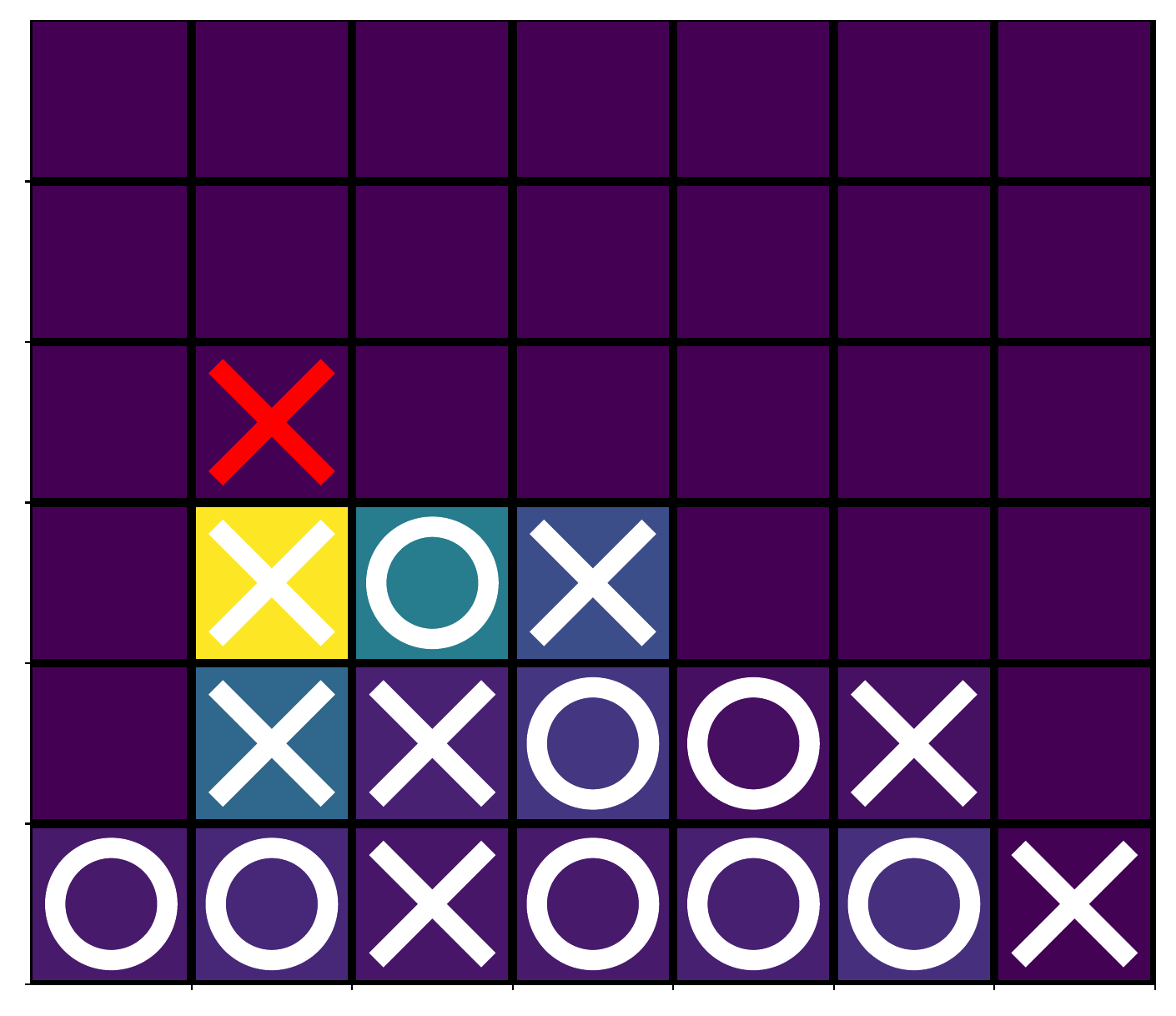}
        &
        \includegraphics[width=0.112\textwidth]{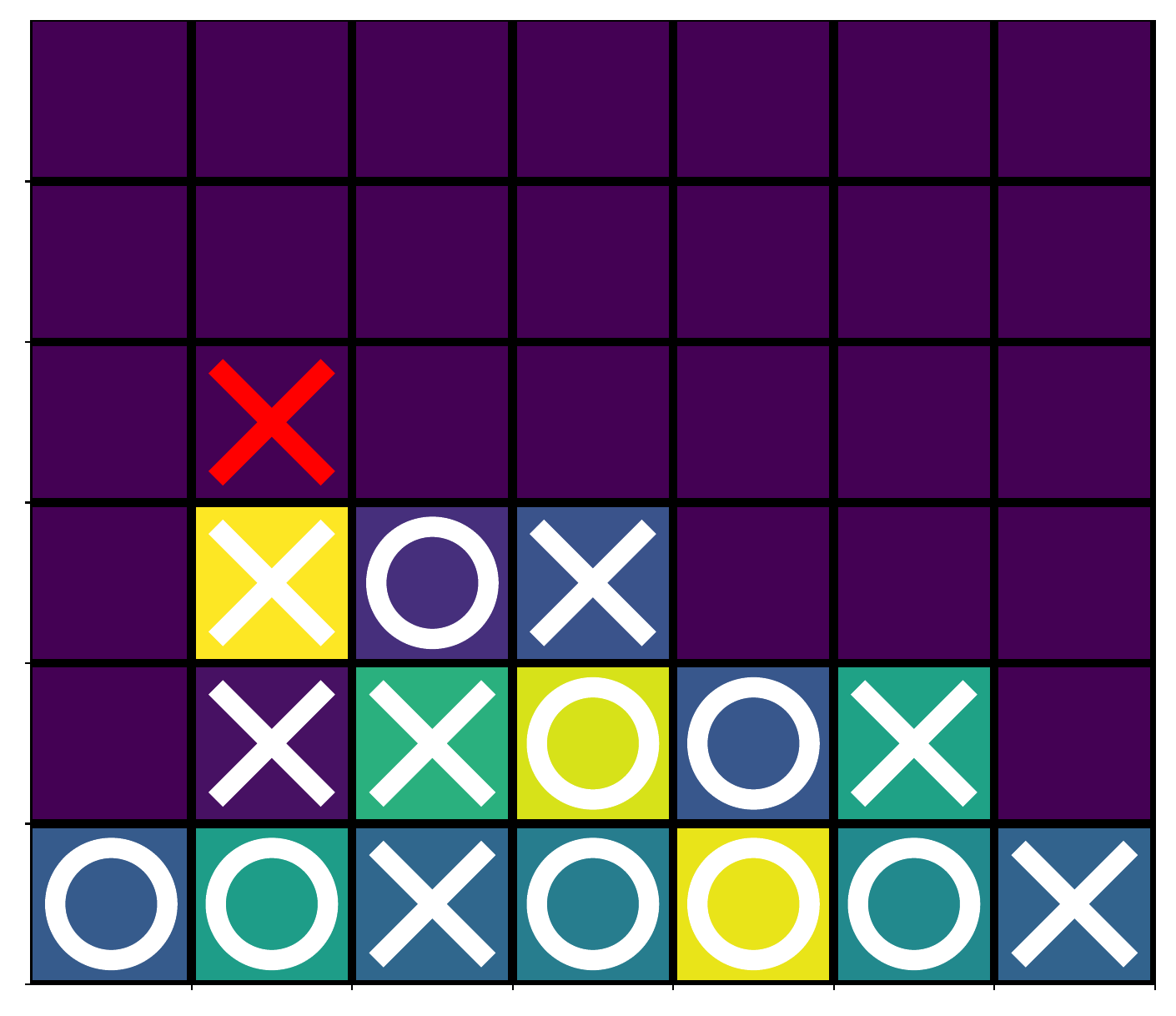}
        &
        \includegraphics[width=0.112\textwidth]{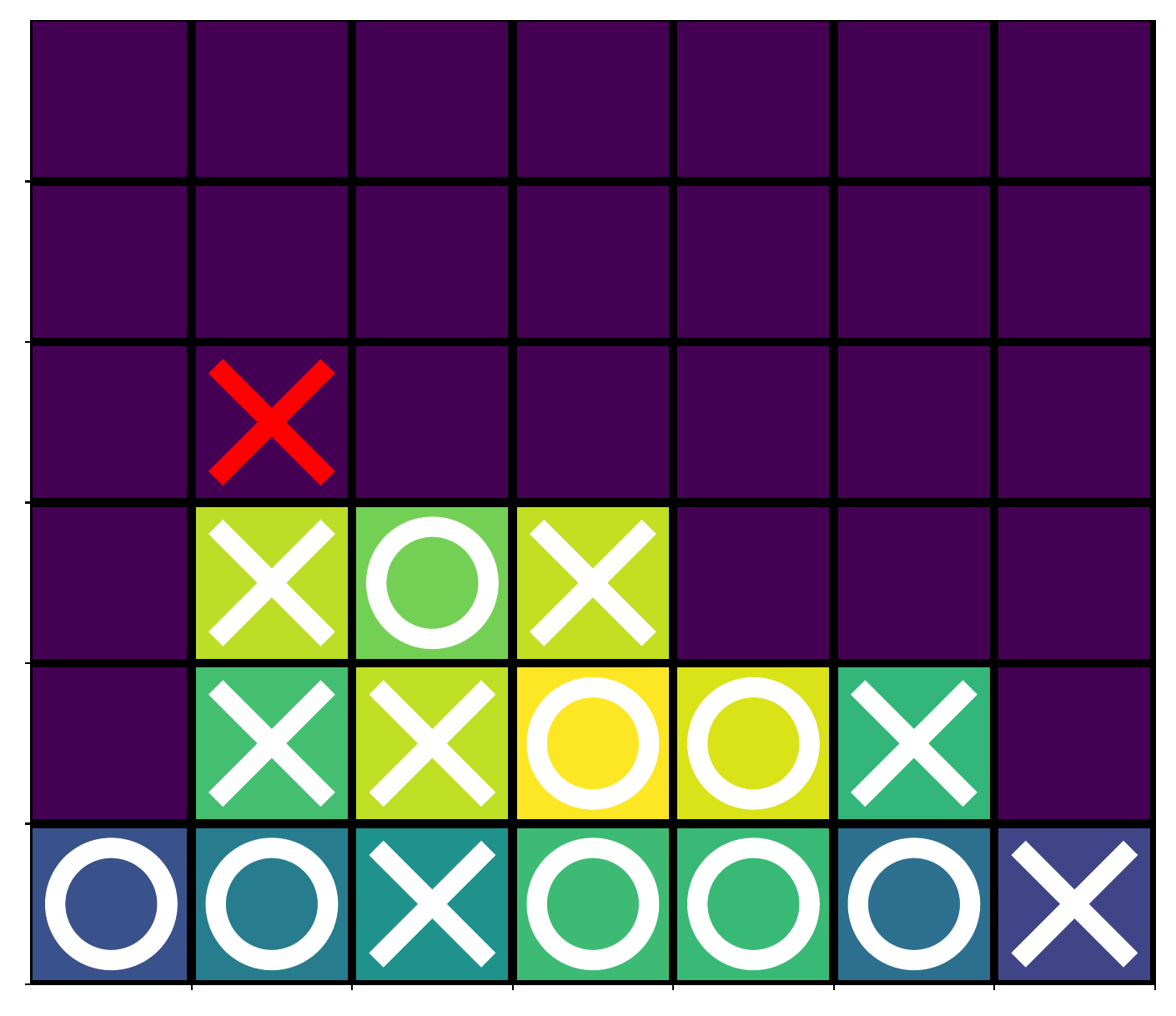}
        &
        \includegraphics[width=0.112\textwidth]{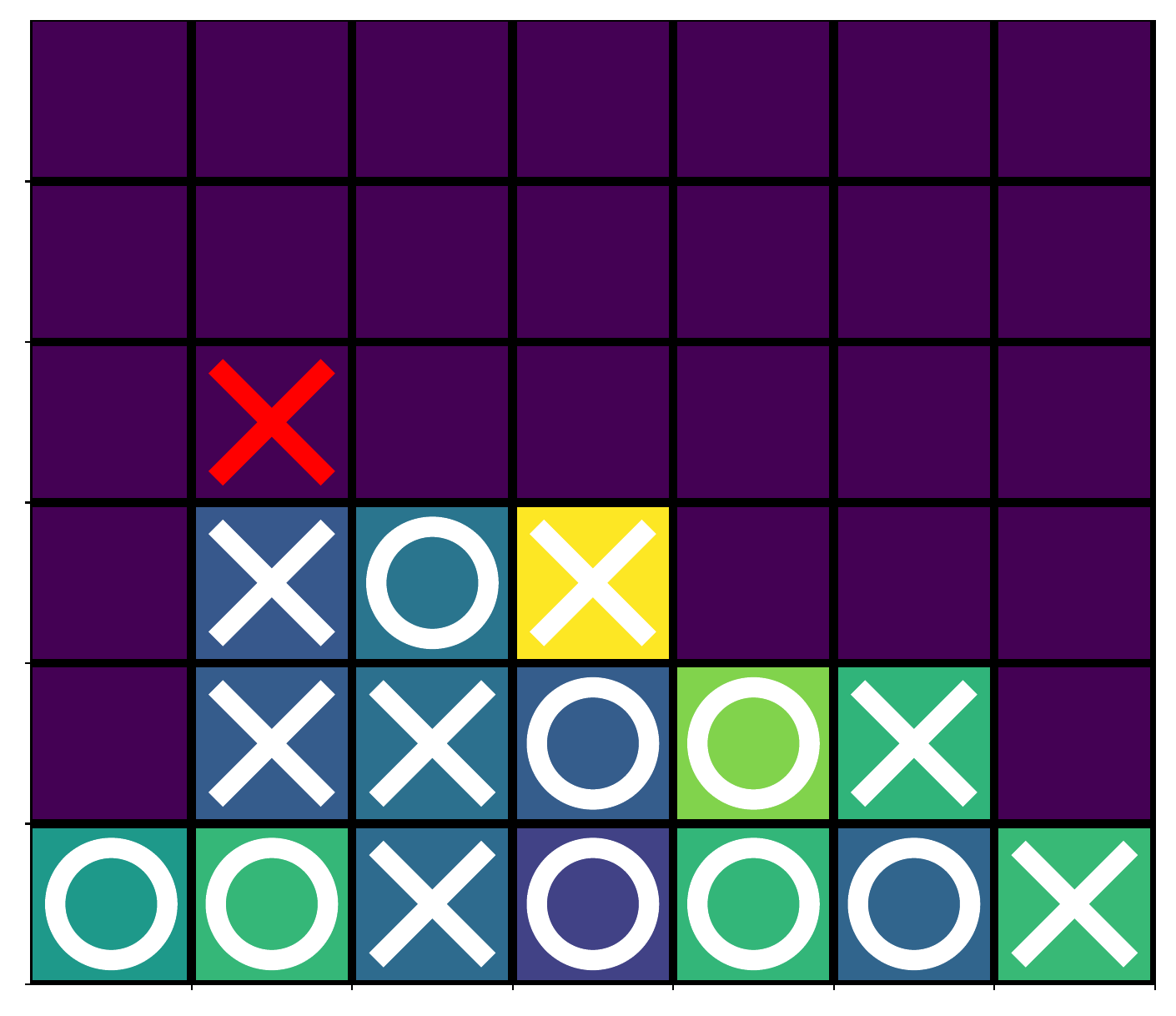}
        &
        \includegraphics[width=0.112\textwidth]{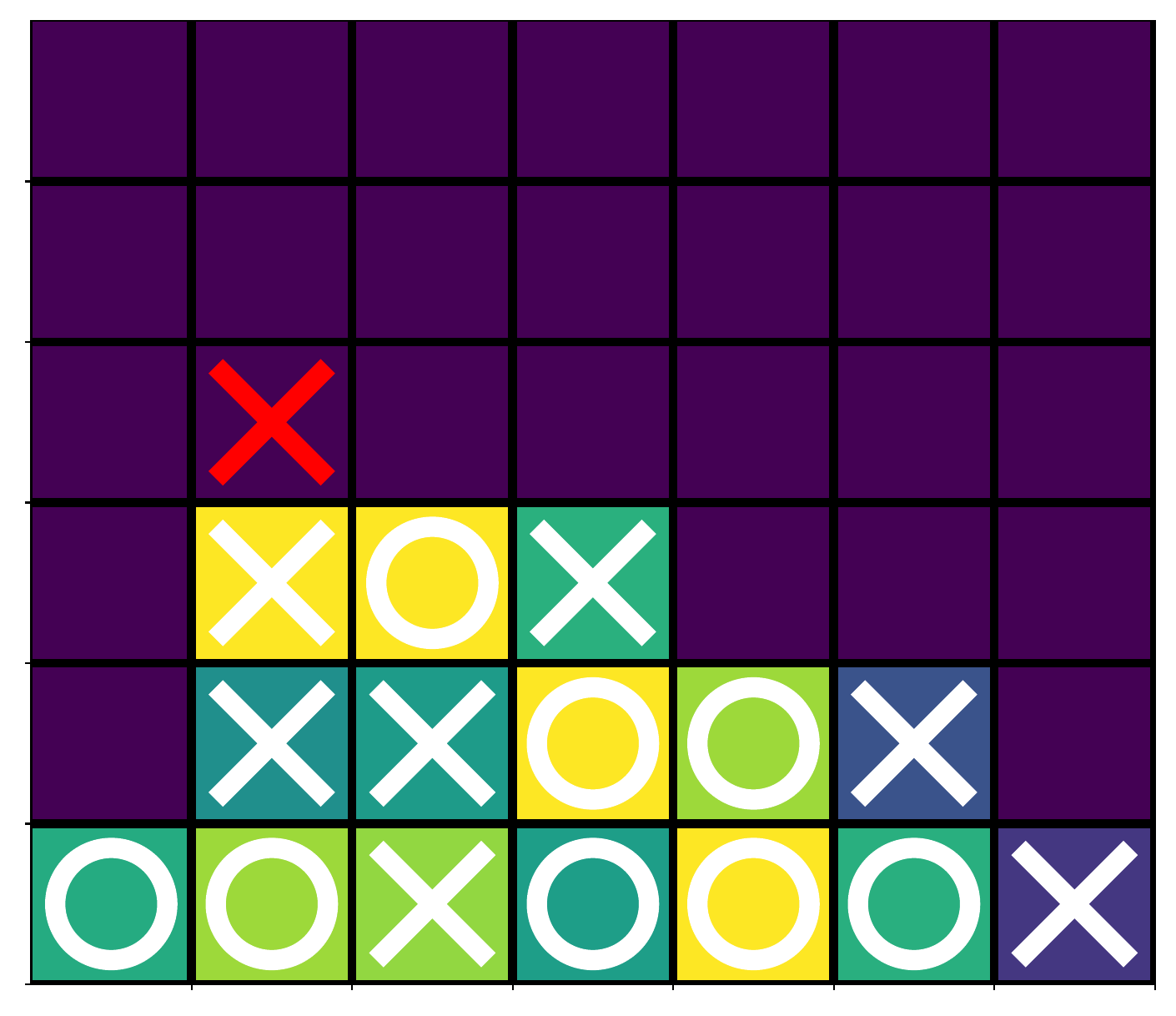}
        &
        \multirow{3}{*}[0.084\textwidth]{
        \includegraphics[width=0.051\textwidth]{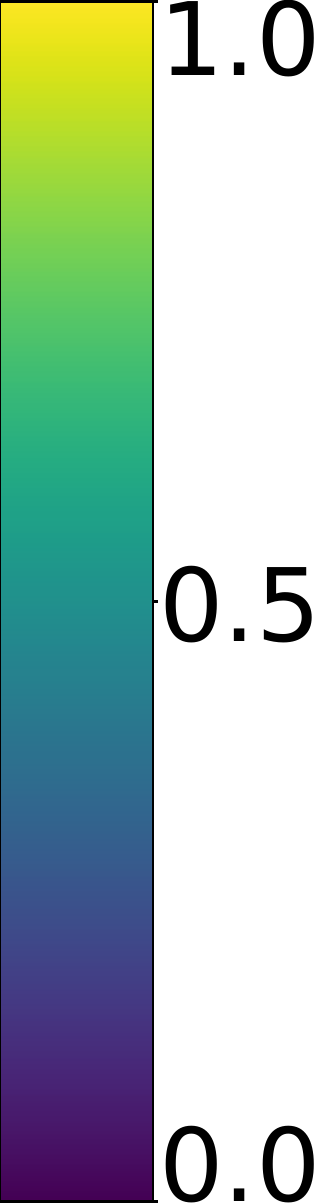}
        }
        \\
        \includegraphics[width=0.112\textwidth]{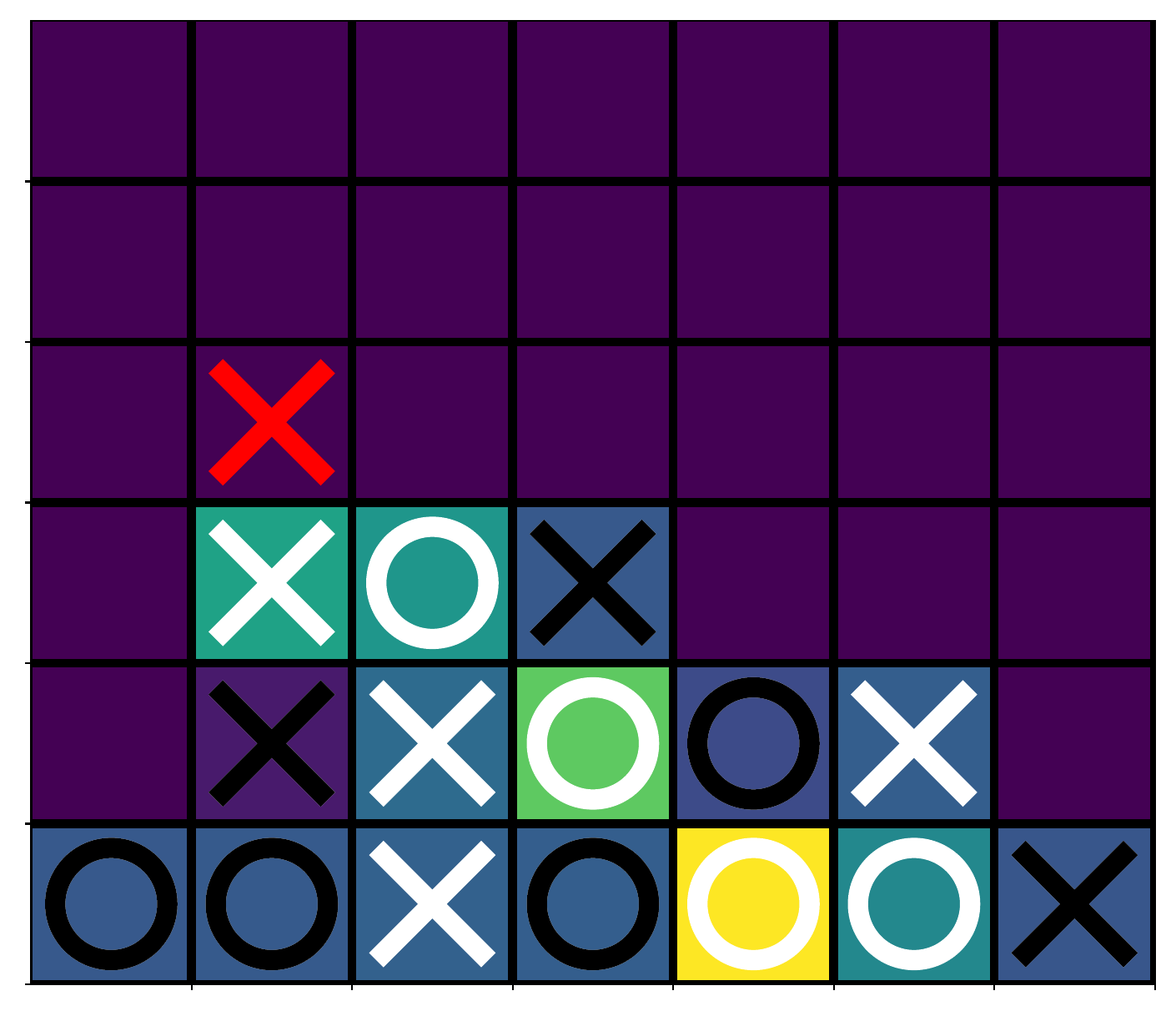}
        &
        \includegraphics[width=0.112\textwidth]{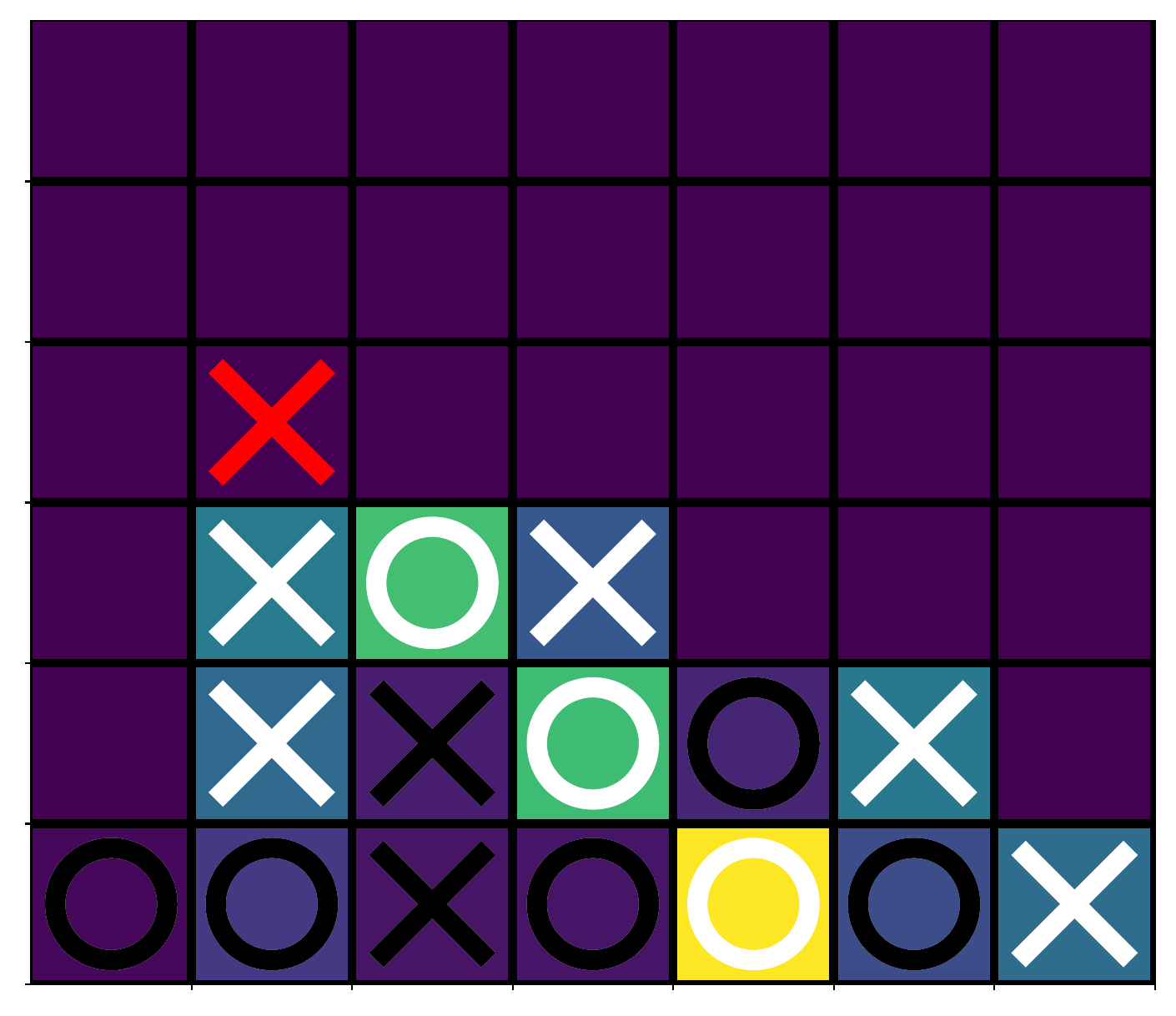}
        &
        \includegraphics[width=0.112\textwidth]{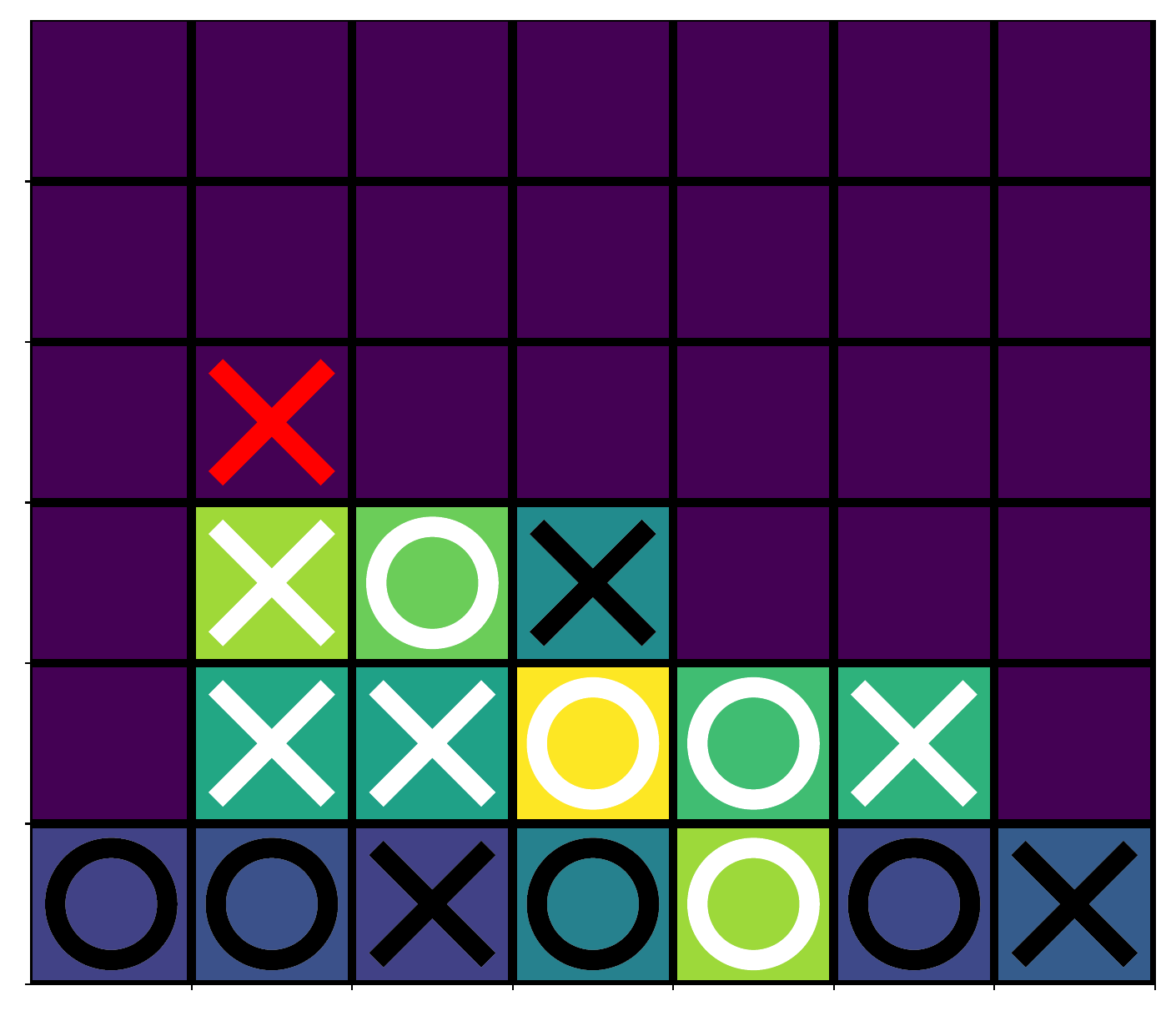}
        &
        \includegraphics[width=0.112\textwidth]{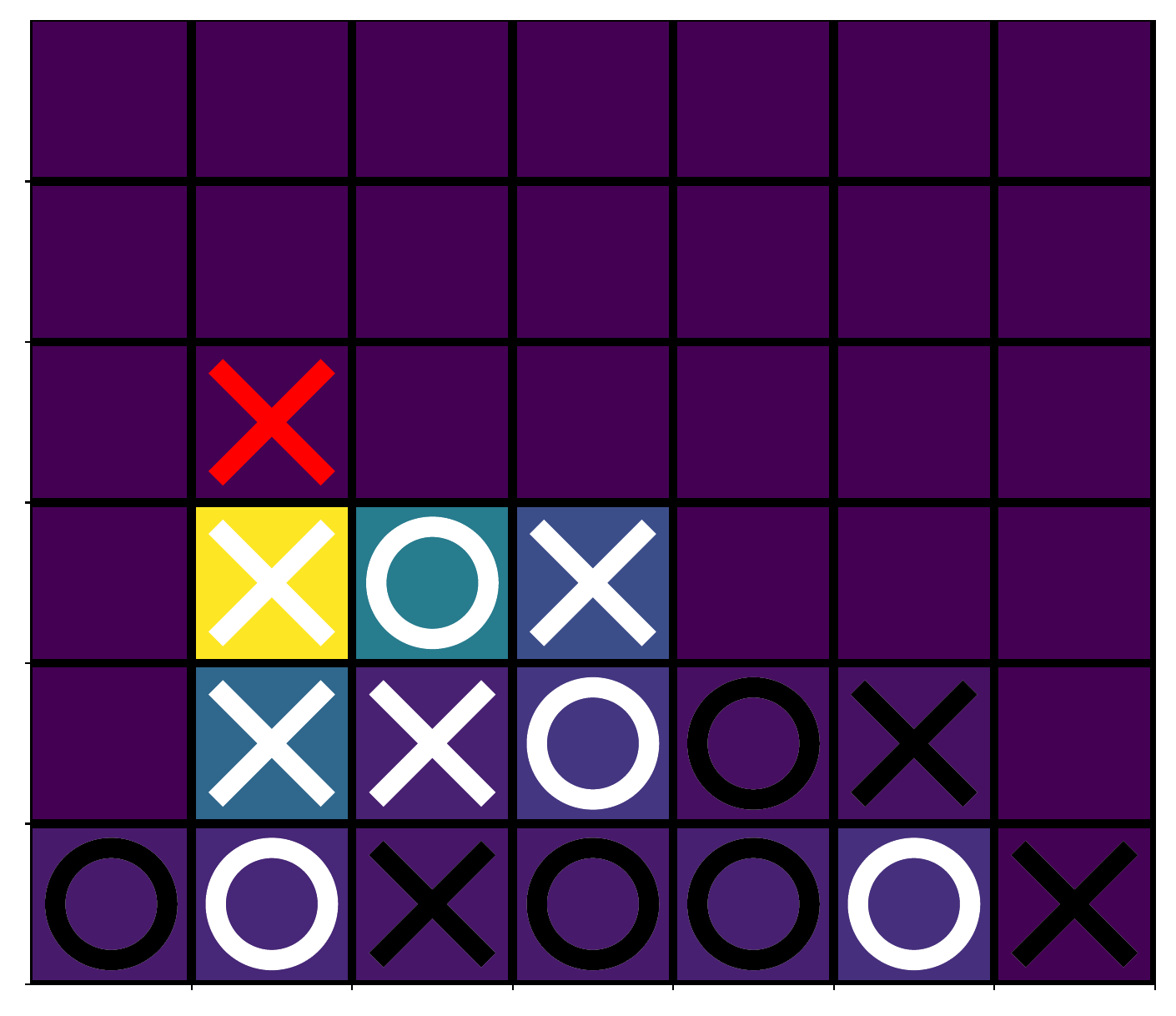}
        &
        \includegraphics[width=0.112\textwidth]{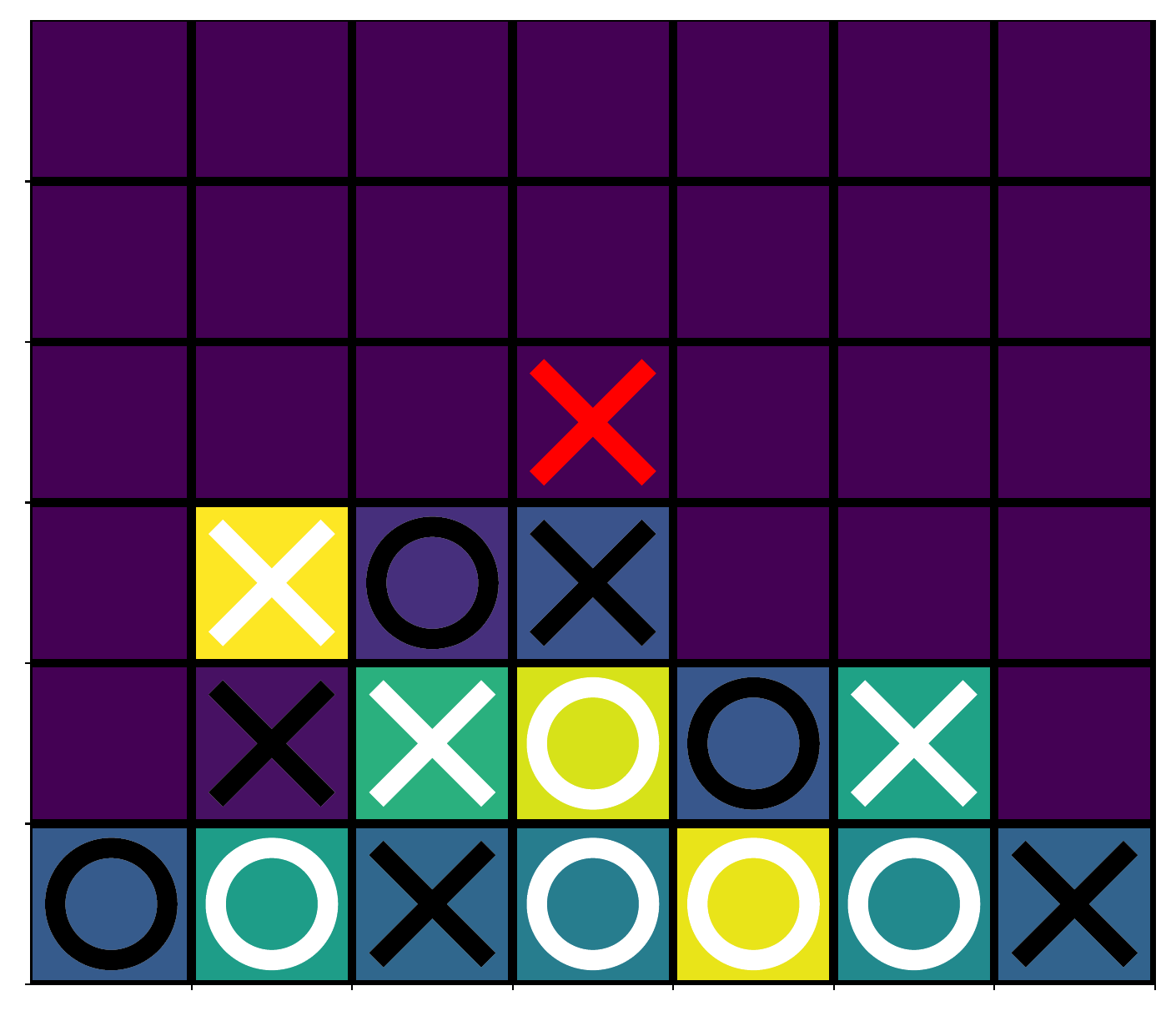}
        &
        \includegraphics[width=0.112\textwidth]{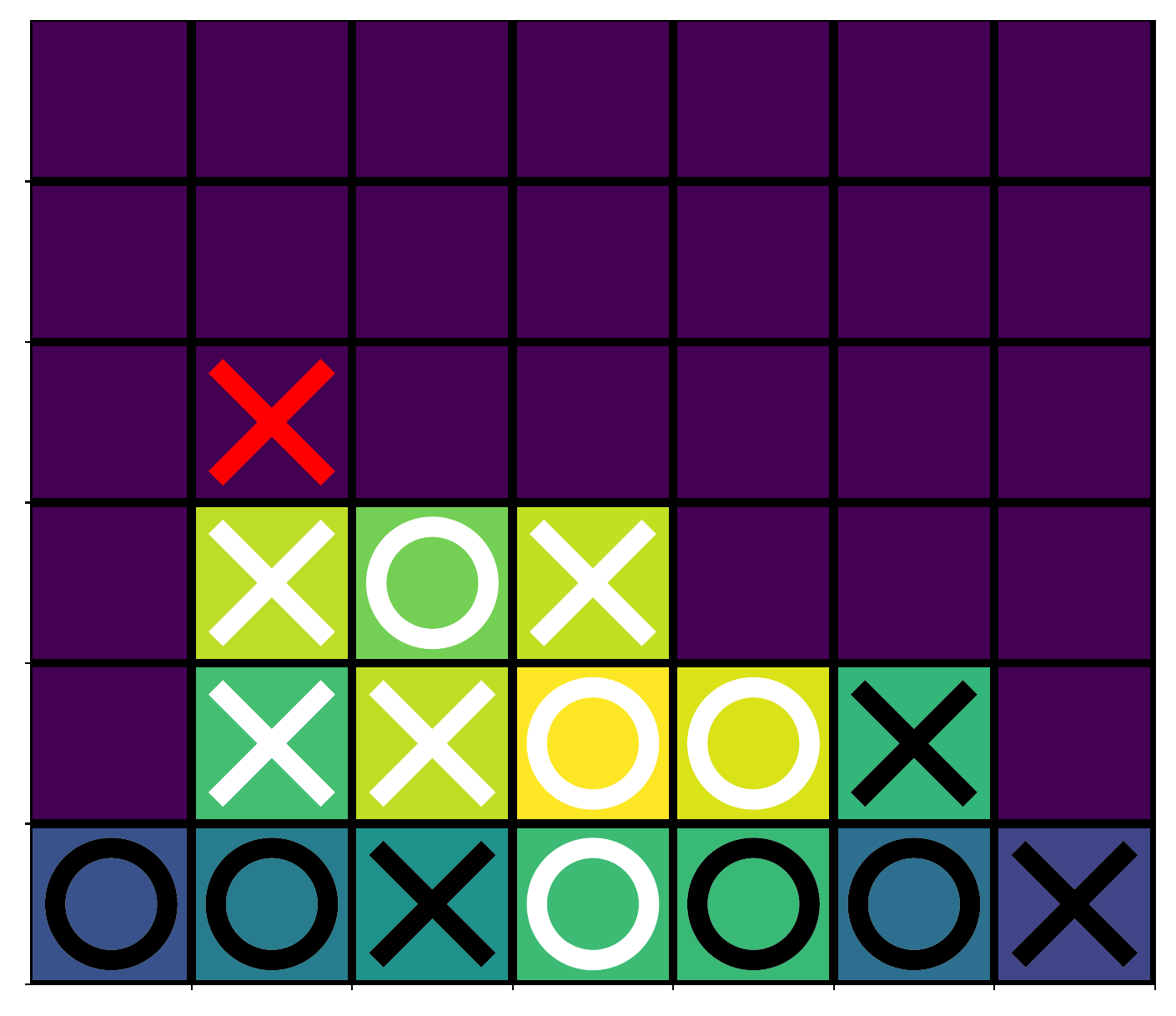}
        &
        \includegraphics[width=0.112\textwidth]{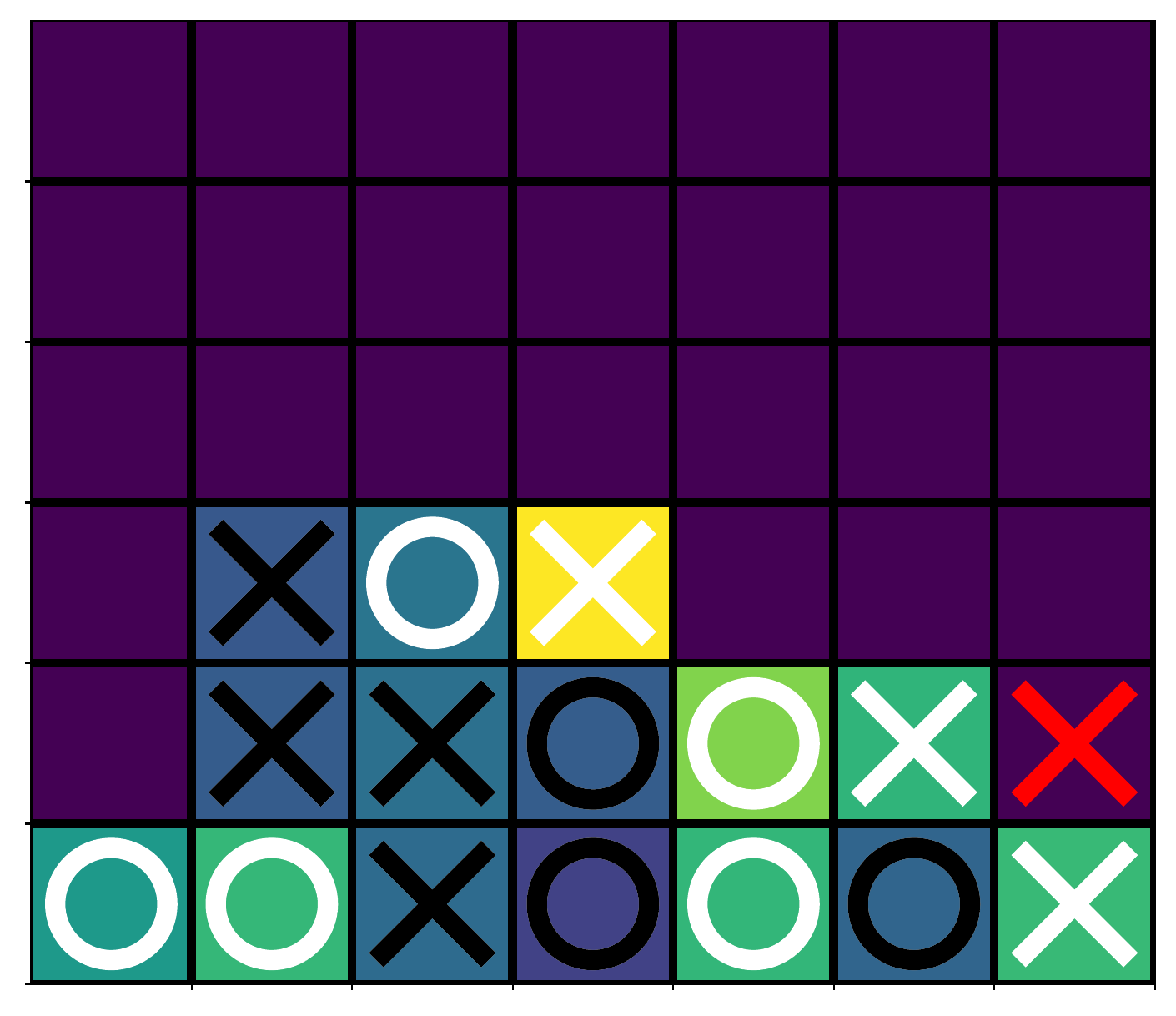}
        &
        \includegraphics[width=0.112\textwidth]{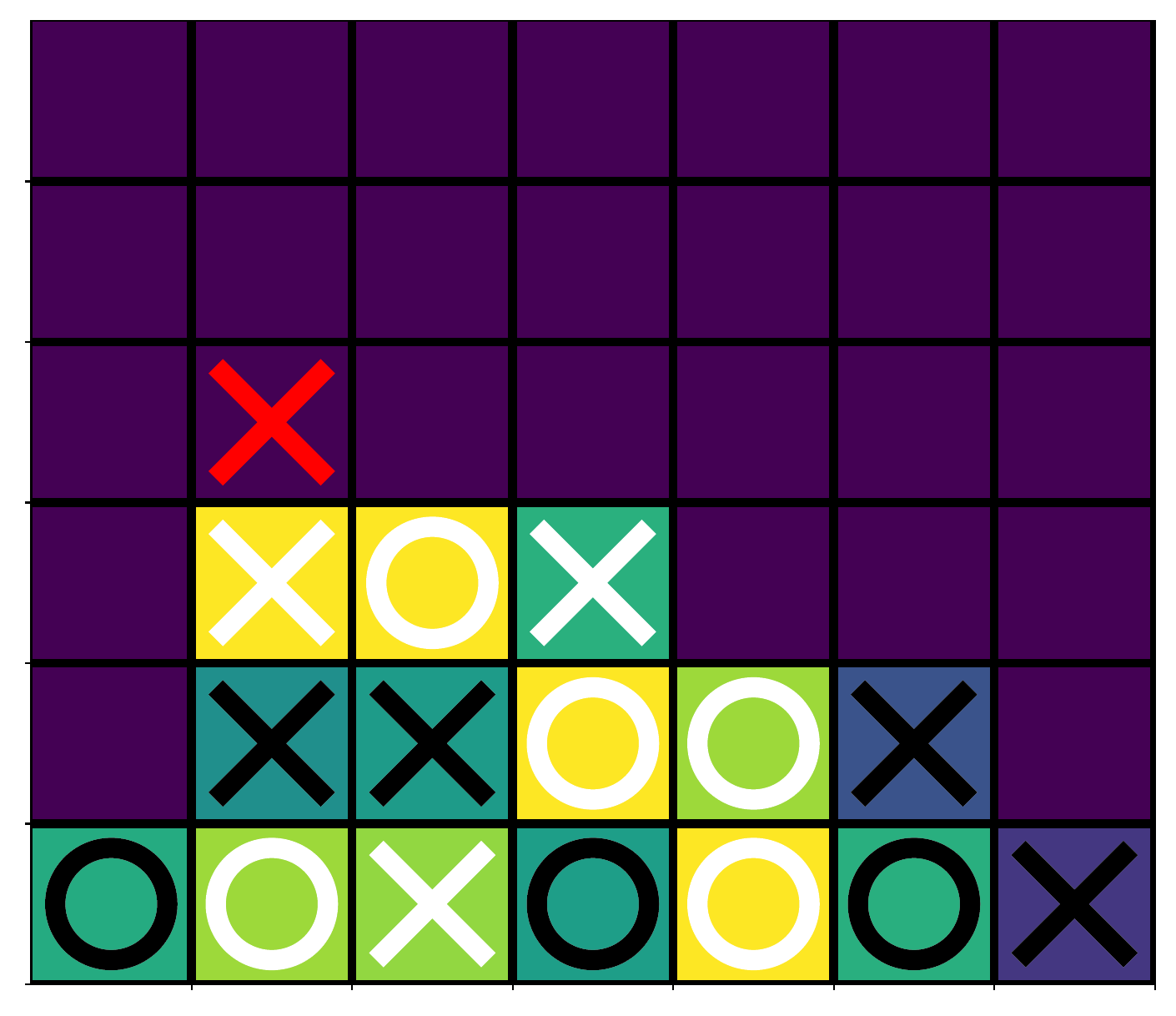}
        &
         \\
         Gradient & DeepShap & GB & SmoothGrad & LRP & DeepTaylor & Random & FW &
    \end{tabular}

\caption{\emph{Top:} One move explained by different saliency maps: The PI-50 agent placed a piece in the second column, blocking a potential win by their opponent (red cross). The saliency maps for this move are shown for each method. \emph{Bottom:} The agent is presented with the colour information of the 50\% most salient pieces according to each method (marked in white). This can lead to a different decision, here for LRP and Random, allowing the opponent to win.}
\label{fig:boards}
\end{figure*}

We can now compare different saliency methods via the setup explained in \cref{fig:highlevel}. As our agent we select PI-50 and allow to show 50\% of colour features, thus remaining on-manifold.
To implement the masker, we let an XAI-method explain the decision of the policy function for the most probable action. For every occupied field we sum the absolute value of saliency score from the first two colour channels (that represent which player occupies the field). The saliency scores on empty fields and on the third channel are ignored. Then we select the 50\% (rounded up) highest scoring colour features and hide the rest from the board state (set them to 0). This state is then used by the player, (PI-50, non-competitive) to select a move. We present an illustration for all used saliency methods in \cref{fig:boards}. Afterwards this process is mirrored by a different saliency method and this is iterated until the game is finished.

We compare the saliency methods Gradient, GuidedBackprop (GB) SmoothGrad, LRP-$\epsilon$, DeepTaylor and Random from the Innvestigate\footnote{\url{https://github.com/albermax/innvestigate}}   \cite{DBLP:journals/corr/abs-1808-04260} and DeepShap from the SHAP\footnote{\url{https://github.com/slundberg/shap}} toolboxes with recommended settings. \emph{Random}, which assigns a random Gaussian noise as saliency value, \emph{Input}, which shows the complete board state, and the FW-method explained in the previous chapter serve as comparison for the other methods. 

\begin{figure*}[t]
    \centering
    \includegraphics[width=\textwidth]{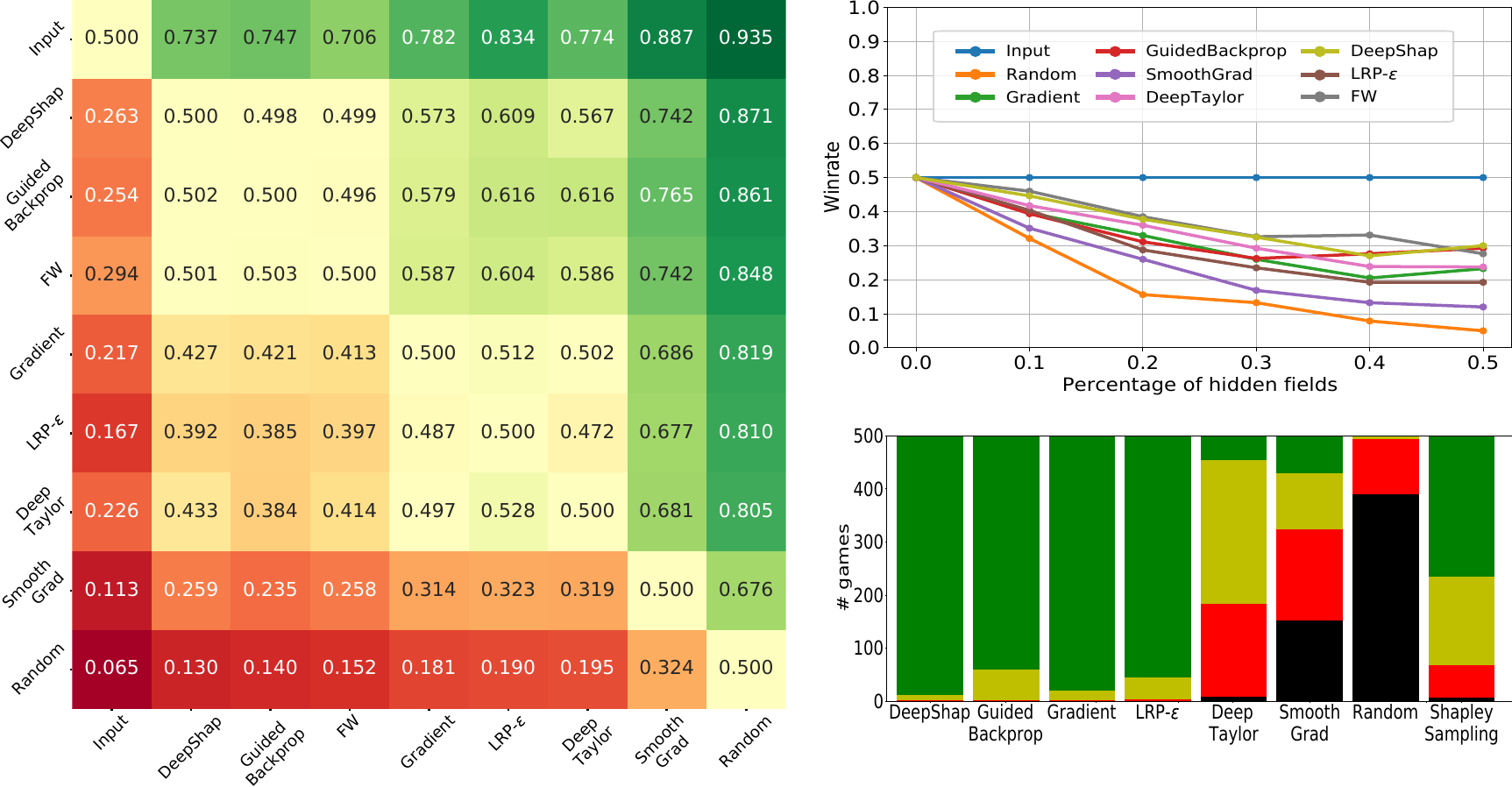}
    \caption{Comparison of different XAI-methods playing Connect Four. \emph{Left:} Win rates row vs column, with draws counting as $\frac{1}{2}$. \emph{Top right:} Win rate of PI-50 with varying percentage of hidden information against itself. The partial colour features have been selected using the XAI-methods. The win rate of the agent decays slower, when the method is better at selecting the crucial information. \emph{Bottom right:} Comparison of the XAI-methods in the ground truth task described in \cref{sec:ground_truth}. We can see that DeepShap and GuidedBackprop perform well in all these tasks.}
    \label{fig:tournament}
\end{figure*}

We let each saliency method compete in a round-robin tournament with 1000 games for each encounter. In case of a draw, both methods score half a victory. We display the number of victories for each encounter in \cref{fig:tournament} together with two challenges described in \cref{sec:ground_truth} and \cref{sec:benchmarking}.

\paragraph{Results} The results are displayed in \cref{fig:tournament}. The methods form two groups of performance. DeepShap, GuidedBackprop and the Frank-Wolfe-based method perform best and equally well. The second group is formed by gradient, LRP and DeepTaylor who show a weaker performance. SmoothGrad has the worst showing, potentially owing to the fact that it does not automatically sample other valid board situations. These results are confirmed by the information-performance graphs, introduced in \cref{sec:benchmarking} for play against PI-50 with full information, where instead of randomly selecting the revealed features they are selected by each saliency method, see \cref{fig:tournament}.

Regarding FW, we showed in \cref{sec:tlight} that the optimiser does not always find what we consider the important pieces but rather the pieces that ensure the policy network makes the right decision. Thus DeepShap and GuidedBackprop compare very favourably considering FW optimises directly for winning the game.
Additionally, DeepShap, Gradient, LRP-$\epsilon$ and GuidedBackprop severely outperform Shapley Sampling in finding the ground-truth pieces. This shows that heuristic methods can have merit over theoretically founded one. We explain a possible reason for the weak performance of the Shapley values and a way forward to improve it in \cref{sec:conclusion}.

\section{Conclusion}\label{sec:conclusion}

We have demonstrated that simple game setups can be used to train agents capable of handling missing features. This allowed us to design a proxy task for XAI-methods based on the idea that such agents make better decisions if provided more relevant information. We evaluate a collection of saliency methods and see strong performances for DeepShap and GuidedBackprop.

As explained in \cref{sec:related_work}, using proxy tasks that evaluate the classifier off-manifold can have paradoxical consequence such as yielding ``superstimulus'' masks as the optimal strategy. To avoid this, we train our agents on the partial information used in our proxy task.
Using the resulting characteristic function given by the policy network, we can directly calculate Shapley values via sampling. This attribution method is theoretically well understood and only relies inputs that were part of the training manifold which justifies trusting the resulting saliency maps.

However, one problem we discovered is that simply extending the training manifold is not enough if the training objective does not strongly regulate behaviour of the classifier.
In our example the policy network is trained to predict sensible game actions---a task that becomes increasingly ill-defined for low information input. The additional entropy term was not strong enough to regulate behaviour towards a uniform distribution over all actions and thus the network output for many low information states was essentially random.
A remedy could lie in using Q-learning instead, even in scenarios where it performs slightly worse than PPO as it only indirectly optimises for policy. Q-learning trains a value function to obey an consistency condition in form of the Bellman-equation \cite{sutton2018reinforcement}. This indirectness can become useful since the objective remains well-defined even if almost no information is given. For a low average loss in terms of the Bellman equation the network would be required to give a conservative estimate of the board value, such as a 50\% chance of winning.  
A similarly option for supervised learning on partial information is to include a default option of ``I don't know'' that is preferred to a wrong answer. Then the objective on low information data becomes well defined and techniques such as Shapley sampling can be used as a theoretically sound saliency method. Tempering with the model to hide biases would require changing on-manifold behaviour which could be detected through performance tests.

In our investigation, the heuristic saliency methods compared very favourably to the more theoretically founded methods. The hope is that future research might prove guaranteed for time efficient methods, e.g. that DeepShap indeed provides a good approximation to the true Shapley values for networks trained on real-world data.
If these saliency attribution methods can be made robust to manipulation, e.g. by approaches from \cite{frye2020shapley} and \cite{anders2020fairwashing}, they could constitute promising tools for XAI.

% For future research, we intend to experiment with different formulations of approximately vanishing such as via the maximum loss as discussed in Section~\ref{section:approximately_vanishing}.
% \newpage

\subsubsection*{Acknowledgements}

 This research was partially supported by the DFG Cluster of Excellence MATH+ (EXC-2046/1, project id 390685689) and the Research Campus Modal funded by the German Federal Ministry of Education and Research (fund numbers 05M14ZAM,05M20ZBM).

\bibliographystyle{apalike}
\bibliography{PPO,introduction,characteristic,bib_icml,related_work}

\newpage
\appendix

\section{Saliency Methods and Off-Manifold Counterfactuals}\label{apx:off_manifold}

Considering a classifier function $f: [0,1]^d \rightarrow [0,1]$ and an input $\bfx \in [0,1]^d$, saliency methods attribute importance (or relevance) values to each input feature $x_i$, with $i\in [d]$, for the classifier decision $f(\bfx)$. In a sense they describe what the classifier focuses on. For a good introduction we refer to \cite{adebayo2018sanity}. We now argue that the three categories of saliency methods introduced in \cref{sec:related_work} all rely on counter-factual inputs that lie off-manifold.

\paragraph{Local Linearisation} For LIME \cite{ribeiro2016should} this is clear since the method samples new inputs $\bfy$ around $\bfx$, labels them $f(\bfy)$ and fits a linear classifier to these new data points.
Arguably, gradient-based methods are always off-manifold for highly non-linear classifiers if the gradient itself is not part of the objective function of the training. In this case, there is principally no reason why the gradient should contain any useful information about the classification. The fact that it often does can be explained for models trained via gradient descent, which implicitly enforces useful gradient information. This however, quickly breaks down when the models are manipulated after training, see \cite{dimanov2020you}. Likewise, for piece-wise constant models that are trained by pseudo-gradients the gradient information is always zero.

\paragraph{Heuristic Backpropagation} For backpropagation-based methods these counterfactual inputs are less obvious. \citeauthor{NIPS2017_7062shap} explain this connection for DeepLift, DeepShap and LRP, which compare the inputs of every layer to baseline values that depends on the specific method \cite{NIPS2017_7062shap}. In this sense they use a counterfactual at every layer instead of only at the input level.

\paragraph{Partial Input} Methods that derive characteristic functions from standard classifiers do this mostly via expectation values \cite{frye2020shapley} over a conditional distribution of counterfactual inputs as in \cref{eq:partial}. However, if the distribution is not modelled correctly, which is difficult for real-world data, it is supported mainly off-manifold. Explanation models that use such a characteristic function, e.g in the form of prime implicants (such as RDE, \cite{macdonald2020explaining} or anchors \cite{ribeiro2018anchors}) or for Shapley values will inherit this flaw, as explained in \cite{frye2020shapley}.

\section{Description of the Training Process}\label{apx:training}

 Our training setup is based on \textbf{Algorithm 1} (``PPO, Actor-Critic Style'') in \cite{schulman2017proximal}. This setup was applied to Connect Four as described in \cite{crespo2020master}, and we adopt most of the hyper-parameters for the training of our agents. 
 
 \paragraph{Network Architecture}
We use a modified version of the architecture proposed by \citeauthor{crespo2020master} with two additional fully connected (FC) layers, described in \cref{fig:architecture}. We changed the input dimension to $3\times6\times7$, representing the fields occupied by the first player (red), the second player (blue) or no one respectively.
The input gets transformed by a series of 4 convolutional layers of filter size $3\times 3$ with stride 1, 512 channels and ReLU activations and zero-padding for the first two layers to keep the board shape of $6\times 7$, which is then reduced to $4\times5$ and $2\times3$ after the last two conv-layers respectively. The resulting tensor is flattened and passed through a series of FCs with ReLU activation, one of shape $3072\times 1024$, one $1024\times 512$ and three $512\times 512$. Then we split the output into the policy head with an FC of size $512\times 7$ and into the value head with size $512\times 1$. The policy head has a softmax activation function, the value head a $\tanh$ activation.

\paragraph{Training Parameters}
 Our PPO-agent plays against itself and for every turn saves state, value output, policy output, reward and an indicator for the last move of the game. We give a reward of 1 for wins, 0 for draws, -1 for losses and -2 for illegal moves. Illegal moves end the game and only the last turn gets saved to memory.
 We make use of a discount factor $\gamma = 0.75$ to propagate back reward to obtain discounted rewards for each state. For clipping the policy loss, we set $\epsilon = 0.2$. The total loss weighs the policy loss with 1.0, the value loss with 0.5 and the entropy loss with 0.01. Every 10 games we update the network parameters with Adam on torch standard settings and a learning rate of $l=0.0001$ for 4 steps.
 
%  Discountfactor = 0,75 ; lr = 0.0001 --> Adam optimizer ; 4 epochs for PPO training; collect data from 10 games--> perform leaningstep on formed batch (omit moves of illegal game); rewards: win =1, draw = 0, loss = -1, illegal=-2; clipping_param = 0.2
 
%  PPO: implementation after https://github.com/nikhilbarhate99/PPO-PyTorch/blob/master/PPO.py , ploss = 1.0, vloss = 0.5 and eloss = 0.01
%  
%  objective: 1.0 * policy_loss - 0.5 * value_loss 0.01 * entropy_loss

\section{Partial Shapley Values}\label{apx:partial}

The Shapley Values are the most established attribution method from cooperative game theory, as they are unique in satisfying the following desirable properties: linearity, symmetry, null player and efficiency \cite{shapley201617}.
They can be defined as a sum over all possible permutations of the set $[d]$ of $d$ players as follows:
\[
 \phi_i(\nu) = \frac{1}{d!} \sum_{\pi \in \Pi([d])} \kl{\nu(P^{\pi}_i \cup  \skl{i}) - \nu(P^{\pi}_i)},
\]
where $\Pi([d])$ is the set of all permutations of $[d]$ and $P_i^{\pi}$ the set of all features that precede $i$ in the order $\pi$.

The size of the coalition $P^{\pi}_i$ corresponds to the number of colour features in our Connect Four board states. In our investigation $\nu$ is based on the policy layer of the PI-50 agent who has only been trained up to 50\% missing information. To avoid off-manifold input we thus want to define partial Shapley values that sample only permutations that ensure a coalition size of at least $p d$ for some $p\in [0,1]$.

To achieve this, we define for every player $i\in [d]$ a set of permutations
\[
\Pi_i^{p} = \skl{\pi \in \Pi([d]) \;\text{s.t.}\; \bkl{P_i^{\pi}} \geq pd},
\]
and define partial Shapley values as
 \begin{equation}\label{eq:partial}
 \phi^p_i(\nu) = \frac{1}{d!} \sum_{\pi \in \Pi_i^{p}} \kl{\nu(P^{\pi}_i \cup  \skl{i}) - \nu(P^{\pi}_i)}.
 \end{equation}
Since the (full) Shapley values are the unique attribution method that fulfil the criteria of symmetry, linearity, null player and efficiency, we loose at least one property. We now show that we retain every property except for efficiency.

\paragraph{Symmetry}
If two players $i,j$ are equivalent, i.e. $\nu(S \cup i) = \nu(S \cup j)$ for all coalitions $S$ that contain neither $i$ nor $j$, then the symmetry property requires $\phi_i(\nu) = \phi_j(\nu)$.

When choosing a different set of permutations $\Pi_i$ for each $i$ then this property holds as long as the collection $\skl{\Pi_i}_{i=1}^d$ is symmetric in the sense that 
$\forall i,j \in [d]:\; \Pi_i = \Pi_j[i \leftrightarrow j]$, where $\Pi_j[i \leftrightarrow j]$ means that we exchange the position of the features $i$ and $j$ for every ordering in $\Pi_j$.
It is easy to see that for our $\skl{\Pi_i^{p}}_{i=1}^d$ this is indeed the case. Thus all terms in \cref{eq:partial} are symmetric between $i$ and $j$ and thus the partial Shapley values are symmetric.

\paragraph{Linearity} Linearity means that $\forall i \in [d]: \phi_i(\nu + \omega) = \phi_(\nu) + \phi_i(\omega)$. Since the expression in \cref{eq:partial} is still linear in $\nu$, the linearity property remains

\paragraph{Null Player} The value $\phi_i(\nu)$ is zero for any null player $i$, and $i$ is a null player if $\nu(S) = \nu(S \cup \skl{i})$ for all coalitions $S$ that do not contain $i$. 
This property is trivially true for the partial Shapley values since all summands are zero for a null player.

\paragraph{Efficiency} The partial Shapley values are not necessarily efficient anymore. Consider for example the characteristic function
\[
 \nu(S) = \begin{cases}
  1 & \bkl{S} \geq pd, \\
  0 & \bkl{S} < pd.
 \end{cases}
\]
In this case for every $i\in [d]$ and $\pi \in \Pi^p_i$ we have $\nu(P^{\pi}_i \cup  \skl{i}) = \nu(P^{\pi}_i) = 1$ and thus $\phi^p_i = 0$.
In that case $\sum_i \phi_i(\nu) = 0 \neq 1 = \nu([d]) - \nu(\varnothing)$, which is required by the efficiency criterion.

\section{The FW-method} \label{apx:FrankWolfe}

To find a small set of colour features that ensure a sensible move from the agent we follow the ideas presented in \cite{macdonald2020explaining} and define a rate-distortion functional over a convex set. Our setup is slightly simplified, since we have an advantage in that we can directly deal with partial input without the need to replace the missing features with random variables from a base distribution.

Let $\bfx \in \skl{0,1}^{3\times6\times7}$ be a state describing the game board defined as $[\bfx^{r}, \bfx^{r}, \bfx^o]$, where $\bfx^{r}, \bfx^{b}, \bfx^o \in \skl{0,1}^{6\times 7}$ indicate the fields occupied by the red and blue player as well as the open fields respectively, as illustrated in \cref{fig:highlevel}. For a continuous mask $\bfm \in [0,1]^{6\times 7}$ that indicates which colour information to show, we define the masked state as $\bfx[\bfm] = \ekl{\bfm \odot \bfx^{r}, \bfm \odot \bfx^{b}, \bfx^o}$, where $\odot$ is element-wise multiplication.
Let $a^* = \argmax P(\,\cdot\,; \bfx[\bfm])$ be the chosen action by the policy layer, then we define the policy distortion with regards to the mask $\bfm$ as
\[
 D^{\text{pol}}(\bfm) = \kl{P(a^*; \bfx) - P(a^*; \bfx[\bfm]) }^2.
\]
For a chosen rate of $k\in \N$ we define the optimal mask $m^*$ as
\[
 \bfm^* = \argmin_{\bfm \in \CB_k^{6 \times 7}} D^{\text{pol}}(\bfm), \qquad \text{where}\qquad  \CB_k^d = \skl{\bfv \in [0,1]^d \,\middle|\, \nkl{\bfv}_1 \leq k }
\] 
is the $k$-sparse polytope (see \cite{pokutta2020deep}) with radius $1$ limited to the positive octant. To optimise the objective we use the solver of \citeauthor{pokutta2020deep} made available at \url{https://github.com/ZIB-IOL/StochasticFrankWolfe} with 50 iterations.
For a given state $\bfs$ and most likely action $a^*$ the FW-method thus returns $\bfm^*$ as a saliency map. Oftentimes, multiple $m_i^*$ converge to 1, so we break ties randomly when selecting the most relevant features according to this method.  

\section{Shapley Sampling for PI-100}\label{apx:tlight}

\begin{figure}[h]
    \centering
    
  \centering
  \begin{tabular}{@{}c@{\;}c@{\;}c@{\;}c@{}}
    % && \Large{Shapley Sampling} & \\[5pt]
     & FI & PI-500 & PI-100\\
    \includegraphics[height=4cm]{./img/ampelplots/scale.pdf} &
    \includegraphics[height=4cm]{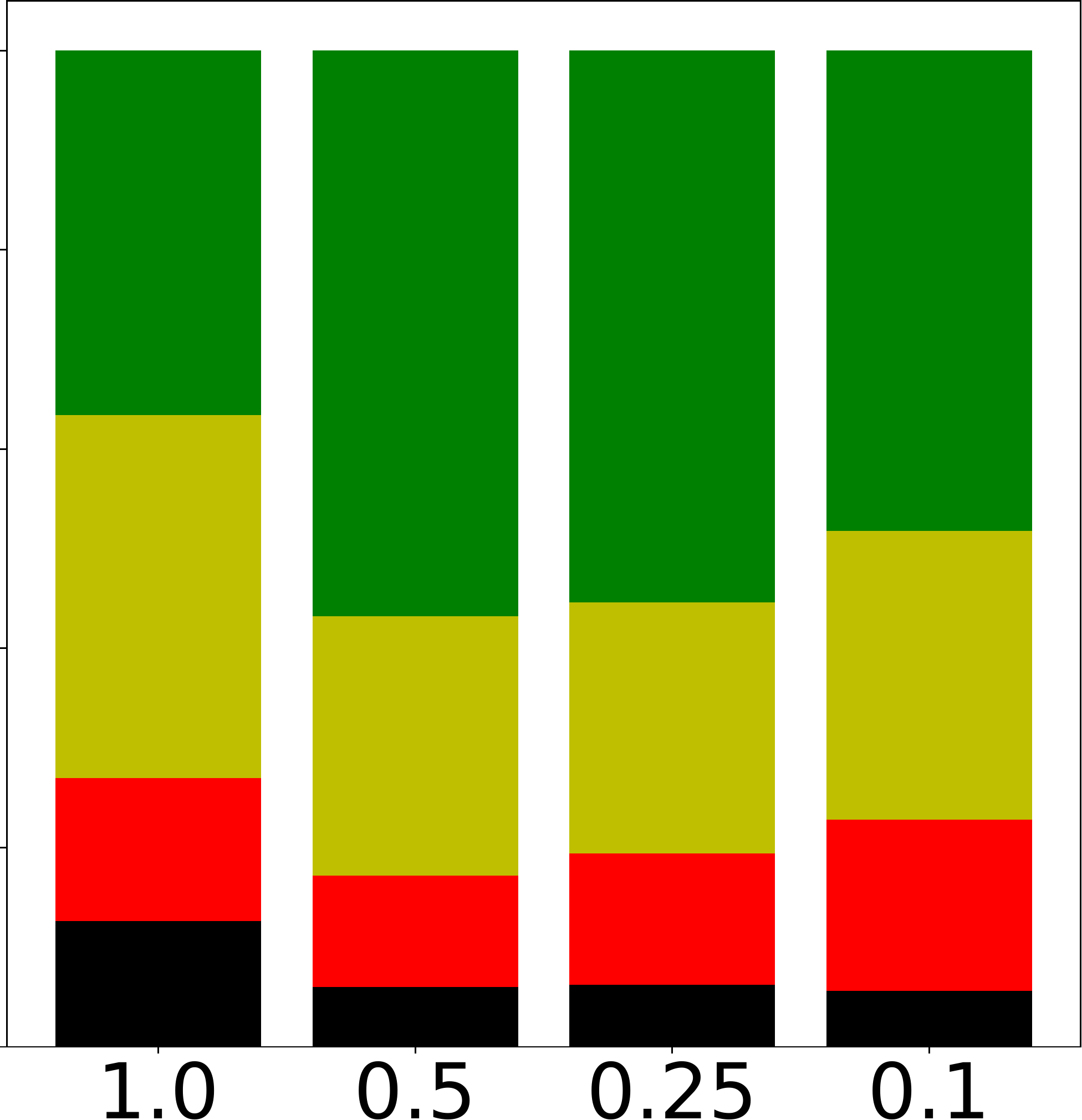} &
    \includegraphics[height=4cm]{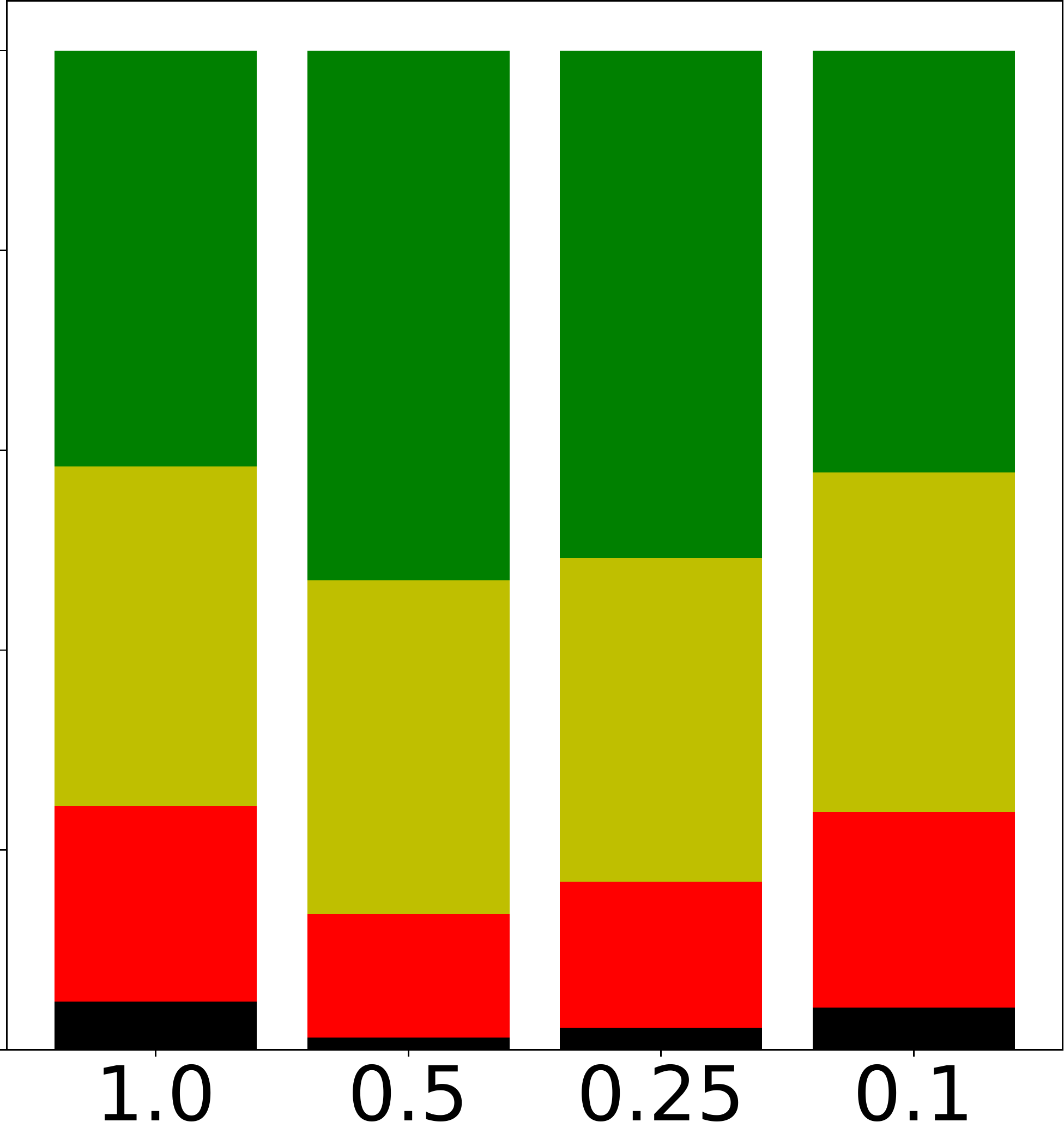} &
    \includegraphics[height=4cm]{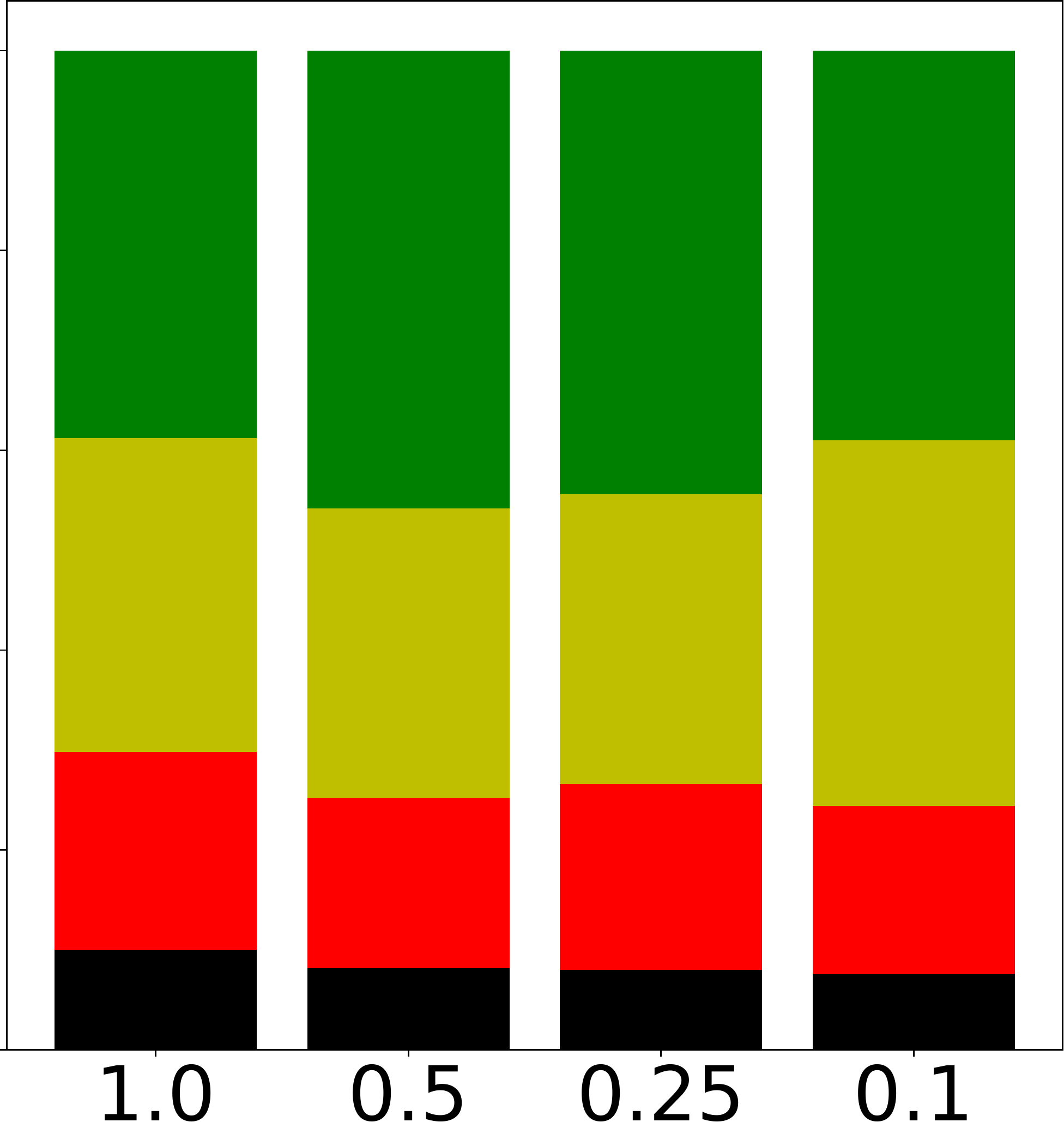}
 \\[-14pt]
    & & $p_h$ & 
  \end{tabular}

    \caption{Comparison of Shapley Sampling for our different agents. The bars show how often the method identified the 3 (green), 2 (yellow), 1 (red) or 0 (black) of the three most important game pieces. The method gives the best results (finding the most pieces) for the PI-50 agent.}
    \label{fig:SHS_PI100}
\end{figure}

We compare the Shapley Sampling approach for the ground-truth task described in \cref{sec:ground_truth} for different agents in \cref{fig:SHS_PI100}. The method works best for the PI-50 agent with $p_h = 0.5$, presumable because it uses the most capable agent with the largest set of permutations that still ensure staying on-manifold. 

\section{Supplementary Tournament Results}

\begin{figure}[t]
    \centering
    \begin{tabular}{cc}
        \includegraphics[width = 0.5\textwidth]{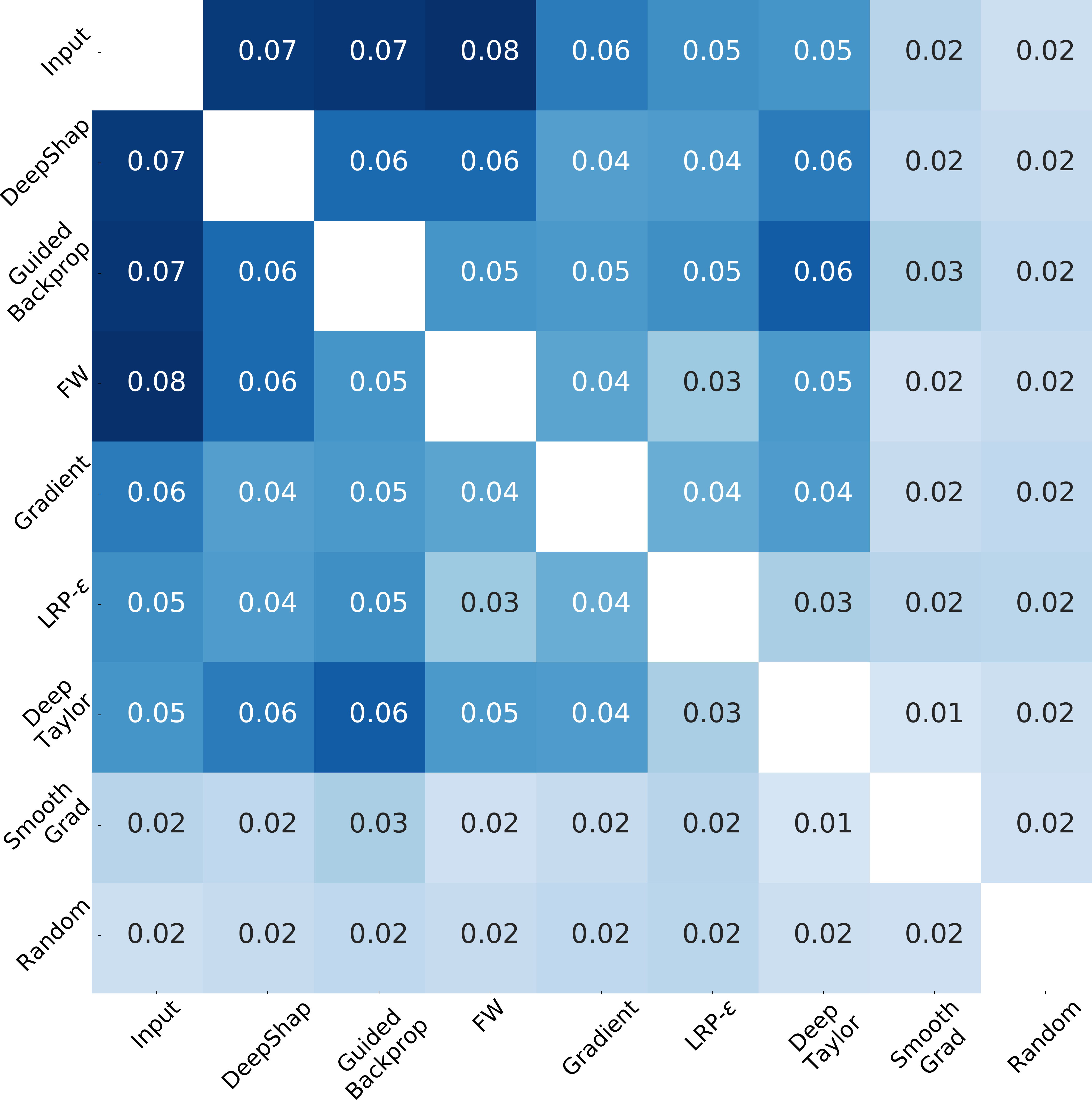}
         & 
         \includegraphics[width = 0.5\textwidth]{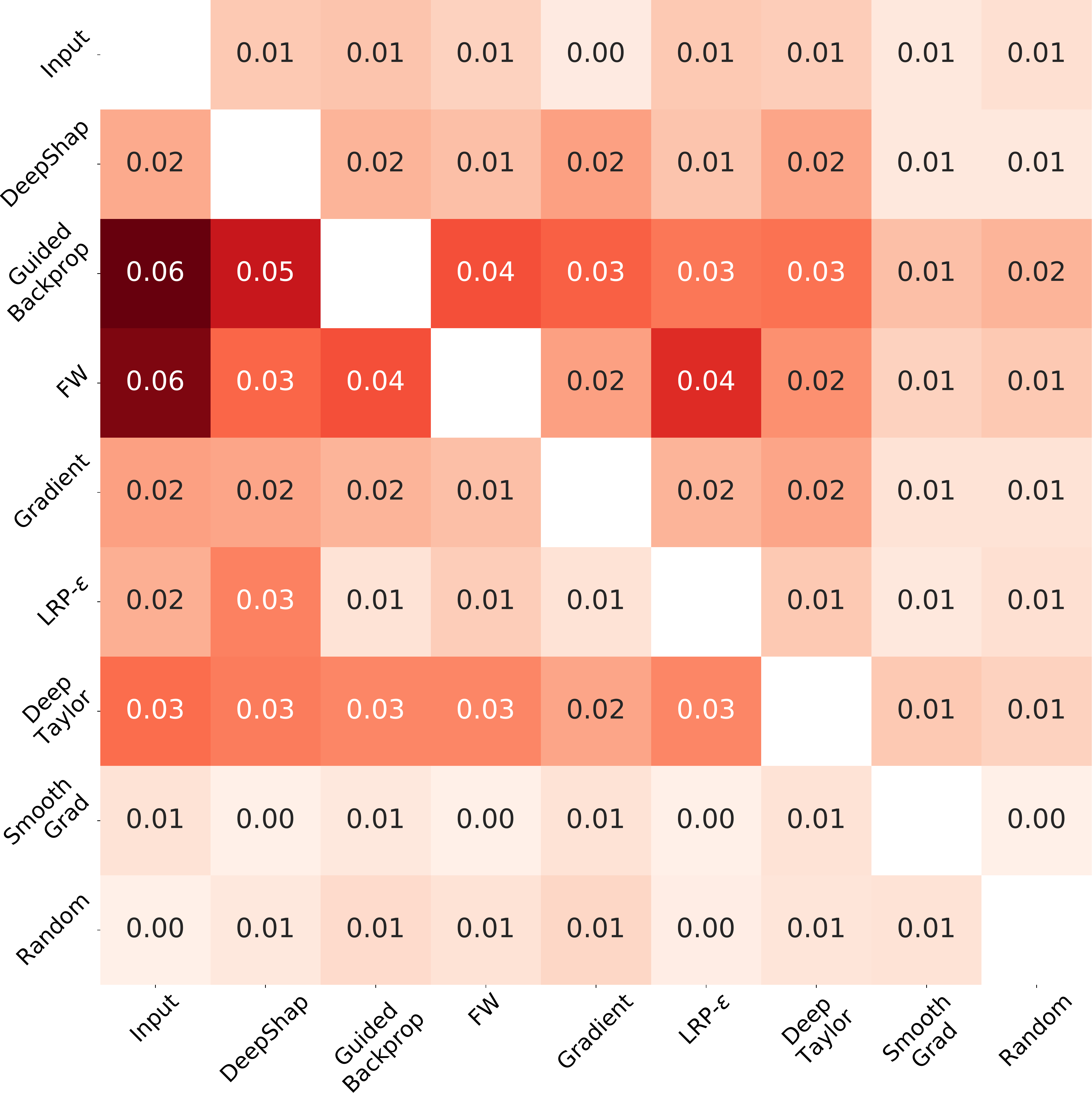}
    \end{tabular}
    \caption{\emph{Left:} Rate of draws for each encounter between saliency methods according to the setup described in \cref{sec:tournament}. \emph{Right:} Rate of illegal moves by either agent for each encounter.}
    \label{fig:tournament_extra}
\end{figure}

We present the number of games of the tournament that ended in draws, as well as the ones where illegal moves were played in \cref{fig:tournament_extra}.

\end{document}